\documentclass[10pt,twocolumn,letterpaper]{article}

\usepackage{iccv}
\iccvfinalcopy

\usepackage{graphicx}
\usepackage{amsmath}
\usepackage{amssymb}
\usepackage{booktabs}
\usepackage{enumitem}
\usepackage{wrapfig}
\usepackage[normalem]{ulem}

\usepackage{times}
\usepackage{epsfig}
\usepackage{graphicx}
\usepackage{amsmath}
\usepackage{amssymb}
\usepackage{booktabs}
\usepackage{xcolor}
\usepackage[pagebackref=true,breaklinks=true,letterpaper=true,colorlinks,bookmarks=false]{hyperref}

\usepackage[linesnumbered,ruled]{algorithm2e}

\usepackage{graphicx}
\usepackage{amsmath, amssymb}
\usepackage{enumitem, array}
\usepackage[flushleft]{threeparttable}
\usepackage{algpseudocode}

\usepackage{xspace}

\makeatletter
\DeclareRobustCommand\onedot{\futurelet\@let@token\@onedot}
\def\@onedot{\ifx\@let@token.\else.\null\fi\xspace}

\makeatother

\usepackage{hyperref}
\hypersetup{hidelinks,
backref=true,
pagebackref=true,
hyperindex=true,
breaklinks=true,
colorlinks=true,
urlcolor=blue,
bookmarks=true,
bookmarksopen=false,
pdftitle={Title},
pdfauthor={Author}}

\usepackage{amssymb}
\usepackage{pifont}
\usepackage{multirow}
\usepackage{wrapfig}

\usepackage{amsmath,amsfonts,bm}

\def\eqref#1{equation~\ref{#1}}

\def\1{\bm{1}}

\DeclareMathAlphabet{\mathsfit}{\encodingdefault}{\sfdefault}{m}{sl}
\SetMathAlphabet{\mathsfit}{bold}{\encodingdefault}{\sfdefault}{bx}{n}

\usepackage{url}
\usepackage{caption}
\usepackage{subcaption}
 \usepackage{multirow}
 \usepackage{enumitem}
 \usepackage{wrapfig}
\usepackage[T1]{fontenc}    
\usepackage{amsfonts}       
\usepackage{nicefrac}       
\usepackage{microtype}      
\usepackage{xcolor}         
\usepackage{float}
\usepackage[accsupp]{axessibility}

\usepackage[capitalize]{cleveref}
\crefname{section}{Sec.}{Secs.}
\Crefname{section}{Section}{Sections}
\Crefname{table}{Table}{Tables}
\crefname{table}{Tab.}{Tabs.}

\def\iccvPaperID{***}

\ificcvfinal\fi

\begin{document}

\title{Controllable One-Shot Face Video Synthesis With Semantic Aware Prior }

\author{Kangning Liu{\thanks{Work done during internship at Google}}$^{*1,2}$, Yu-Chuan Su$^{2}$,  Wei (Alex) Hong$^{2}$,
  {Ruijin Cang $^{\dagger 2}$, Xuhui Jia{\thanks{Joint last author }}$^{\dagger 2}$} \\
  $^1$NYU Center for Data Science   \space  \space $^2$ Google
  }
\maketitle
\ificcvfinal\thispagestyle{empty}\fi
\begin{abstract}
The one-shot talking-head synthesis task aims to animate a source image to another pose and expression, which is dictated by a driving frame. Recent methods rely on warping the appearance feature extracted from the source, by using motion fields estimated from the sparse keypoints, that are learned in an unsupervised manner. Due to their lightweight formulation, they are suitable for video conferencing with reduced bandwidth. However, based on our study, current methods suffer from two major limitations: 1) unsatisfactory generation quality in the case of large head poses and the existence of observable pose misalignment between the source and the first frame in driving videos. 2) fail to capture fine yet critical face motion details due to the lack of semantic understanding and appropriate face geometry regularization. 
To address these shortcomings, we propose a novel method that leverages the rich face prior information, the proposed model can generate face videos with improved semantic consistency (improve baseline by $7\%$ in average keypoint distance) and expression-preserving (outperform baseline by $15 \%$ in average emotion embedding distance) under equivalent bandwidth. Additionally, incorporating such prior information provides us with a convenient interface to achieve highly controllable generation in terms of both pose and expression.

\end{abstract}

\section{Introduction}
\label{sec:intro}

The ability to animate static images has numerous important applications such as film production, digital advertisement, video dubbing and immersive communication~\cite{deng2008computer,sha2021deep}. In this paper, we revisit the problem of one-shot face video synthesis, where a portrait image is given, and animated by a driving video. The resulting video, therefore, preserves the face appearance from the portrait image, while pose and expression are inherited from the driving video. 

There are many existing attempts to tackle face video synthesis by using deep generative models, and they can be broadly summarized into two paradigms. The first line of works~\cite{kim2018deep,koujan2020head2head,doukas2021head2head++,papantoniou2022neural} relies solely on learning-based image reconstruction, which typically is subject-specific and requires hours of video footage of source subjects to fit the appearance and motion of a source, so that in the inference, the model can map the source directly to the target video in an image-to-image translation manner. Recently, various methods ~\cite{yao2020mesh,doukas2021headgan,wang2019few,ha2020marionette,zakharov2020fast,siarohin2019first,wang2021one} have been proposed to build a generic model that is applicable to any source subject under one-shot setting, which is much more challenging given the minimal input. Thanks to the advanced power of GANs and the disentangled representation in appearance and motion, keypoints-based motion-warping methods have made notable progress in driving expression and pose in arbitrary view. Specifically, Monkey-Net~\cite{siarohin2019animating} proposes a network that transfers the deformation from sparse to dense motion flow. FOMM~\cite{siarohin2019first} extends Monkey-Net via the first-order local transformations. Then, Face-vid2vid~\cite{wang2021one} improves FOMM via a learned 3D unsupervised keypoints for free-view talking head generation. Besides achieving good photo-realism, their formulation remains lightweight, only involving a single source image and a sequence of compact self-learned keypoints, which essentially enables some novel applications like low bandwidth video conferencing.

\begin{figure}[t]
\centering
\resizebox{1\linewidth}{!}{
    \setlength{\fboxrule}{4pt} 
    \setlength{\fboxsep}{0cm}
    \begin{tabular}{c c c}
        Source & Face-vid2vid & Driving\\
        \includegraphics[width=0.4\linewidth]{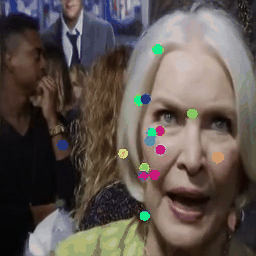} &
        \includegraphics[width=0.4\linewidth]{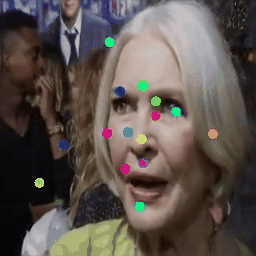} &
        \includegraphics[width=0.4\linewidth]{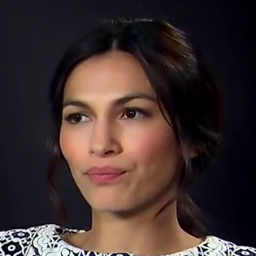}\\
        \includegraphics[width=0.4\linewidth]{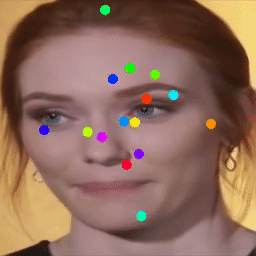} &
        \includegraphics[width=0.4\linewidth]{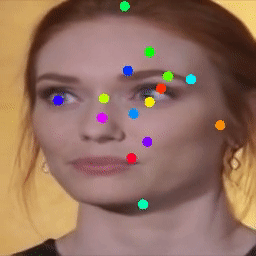} &
        \includegraphics[width=0.4\linewidth]{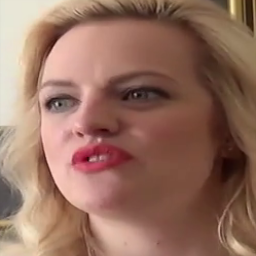}\\
    \end{tabular}  
}

\vspace{-8pt}
\caption{Problems of existing keypoint-based talking-head method. Despite it's superior performance and compact representation, we observe two common problems stem from the keypoint design. First, the keypoints may fail to establish correct correspondence between source and driving frame especially when the head poses are significantly different, which leads to incorrect reconstruction. Second, the keypoints are not dense enough around semantically important regions like the mouth, so the model fails to transfer fine-grained expression or motion.}
\label{fig:compare_fig_val_fail}
\vspace{-12pt}
\end{figure}

However, based on our studies, these end-to-end learned unsupervised keypoints have two major limitations as can be seen in Fig.~\ref{fig:compare_fig_val_fail}. Firstly, unsatisfactory generation quality in the case of large head poses, and the existence of observable pose misalignment between the source and the first frame in driving videos. We conjecture the former is highly correlated with the missing of external supervision on the keypoints learning. While the latter hinders their usability, as such assumption is not always met therefore it's not guaranteed that generated samples would follow the driver. Secondly, failing to capture fine yet critical facial motion details, e.g., in row 2, the driver frame has a smile-like expression, however, its magnitude can not meticulously be captured in the generated frame. We argue this is due to the lack of semantic understanding and appropriate face geometry regularization. We also visualized the learned unsupervised keypoints in Fig.~\ref{fig:compare_fig_val_fail}, intuitively, limited keypoints are assigned to the important regions such as the eyes and mouth. Lastly, we observe that the two aforementioned issues can not be simply solved via scaling up the number of unsupervised keypoints. Adding more keypoints usually does not lead to a more accurate flow estimation, on the contrary, it may even deteriorate the performance.

We propose to improve the motion-warping-based methods by leveraging a readily available prior knowledge of faces. As different from other objects, face, a regular object, has been well studied since the early graphics community, and 3D faces model provides a powerful prior for rendering and editing via the parameters modulation. Therefore, we simply grab an off-the-shelf face model: 3DMM~\cite{blanz1999morphable}, and plug it into our pipeline, but any other face model shall flexibly work. We focus on identifying the best generative strategy compatible with low bandwidth cost and real-time inference. Different from the previous methods ~\cite{yao2020mesh,doukas2021headgan} that rely on dense mesh, we demonstrate that by only resorting to a few sparse keypoints extracted from prior, our learned generator can achieve identity preservation, and better transfer the pose and expression even under the large head poses.

Our contributions can be summarized in the followings: 1) We empirically verify the limitation of current unsupervised keypoint-based one-shot face video synthesis methods, and propose a novel framework equipped with a plug-and-play face prior, besides robustness, it enables several new functionalities: flexible expression editing and novel view synthesis. 2) We provide a comparative analysis of design choices to address the flaw that leads to inaccurate warping. Moreover, person-agnostic expression information is utilized to further enhance the model performance in faithfully retaining fine-grained expression. 3) We have done comprehensive experiments to investigate the trade-off between synthesis quality (geometric correspondences, expression preserving) and bandwidth, our strategy consistently outperforms the baseline model across different bandwidths.

\section{Related Work}
\label{sec:related}

\noindent \textbf{Keypoint-based warping method} 
Keypoint-based warping models represent the pose and expression transformations using explicit motion fields estimated via keypoints. These model then warp and synthesize the target faces based on the estimated motion fields.
Early models use facial landmarks as such keypoints~\cite{averbuch2017bringing, geng2018warp} and warp directly on the source image,  which often leads to unnatural head deformation.
As facial landmarks allow identity-related information from the driver to be transferred into the generated video, most methods are limited in achieving identity preserving~\cite{geng2018warp,averbuch2017bringing,zakharov2020fast}.
 MarioNETte~\cite{ha2020marionette} tries to solve this problem with a landmark formation method.  However, the result is still not satisfactory for the one-shot case~\cite{yao2020mesh}. 

Recently, unsupervised keypoints based warping methods, which leverage first-order motion to warp the latent feature maps, achieve state-of-the-art performance with much more compact keypoints than landmarks~\cite{siarohin2019animating, siarohin2019first,siarohin2021motion,wang2021one, hong2022depth, mallya2022implicit, zhao2022thin,tao2022structure}.
Face-vid2vid~\cite{wang2021one} proposes to detect unsupervised 3D keypoints in the canonical space, which disengages identity from expression and obtains improved identity-preserving ability. Furthermore,  conducting warping in 3D space allows free-view control. 
Following the unsupervised keypoint-based paradigm, DaGAN~\cite{hong2022depth} learns a self-supervised learned depth map as a supplementary modality. Mallya et.~al.~\cite{mallya2022implicit} propose a transformer-based implicit warping method to learn from multiple source images. These techniques are orthogonal to our method and may be combined for better performance.

However, the unsupervised learned keypoints do not necessarily match the precise locations of the face like facial landmarks do~\cite{ren2021pirenderer}. It is tempting to see how landmarks perform when equipped with the more recent flow-estimation method used in the unsupervised keypoint-based framework.  Oquab et.al.~\cite{oquab2021low} is the first to compare the off-the-shelf 2D facial landmarks and the unsupervised learned landmarks under a unified keypoint-based framework~\cite{siarohin2019first}.  They use 2D landmark prediction model to obtain the keypoints, which suffer from the identity loss issue as mentioned before. They reach the conclusion that landmarks are not superior to unsupervised keypoints. Building upon the recent unsupervised 3D keypoint-based methods, we leverage the external semantic-aware prior and achieve improvements across different bandwidth budgets. 

\noindent \textbf{3D-face-based warping method} 3D face reconstruction models~\cite{blanz1999morphable,Feng:SIGGRAPH:2021} provide valuable prior information about human face. Recent methods apply it to 2D images for rendering 3D dense meshes, which are used to learn a dense flow estimator to warp the image in the input space~\cite{yao2020mesh,doukas2021headgan,zhang2021flow}. The warped image is then refined by another generator to create a more realistic image.
Alternatively, Ren et.al. ~\cite{ren2021pirenderer} directly induce the 3D face parameters into a flow estimator without utilizing the motion descriptors (e.g. keypoints, landmarks or edges). However, this leads to spatial information loss~\cite{xu2022designing}, which makes their method achieve worse geometric correspondence than the recent keypoint warping-based methods~\cite{zhong2022geometry,zengfnevr}. In this paper, we mainly leverage the 3D prior to improve the keypoints-based methods.

\noindent \textbf{Direct synthesis methods}  Direct synthesis methods aim to directly generate the result without explicit warping. Early methods~\cite{zakharov2019few,meshry2021learned} take landmarks as the input for guiding generation. Recent studies focus on the extraction and disentanglement of pose and identity in the latent space~\cite{burkov2020neural,liang2022expressive,shu2022few}, where a meta-model is finetuned on a specific person under a few-shot setting. This line of methods enjoys better robustness against large pose changes, but the overall fidelity of the synthesized faces tends to be lower than those produced by warping-based methods~\cite{doukas2021headgan}. In this work, we induce expression code derived from the semantic aware prior to the generator. Our method thus combines the advantage of the warping-based method and the direct synthesis method.

\noindent \textbf{Person-specific methods}  Person-specific methods conduct learning-based reconstruction~\cite{kim2018deep,koujan2020head2head,doukas2021head2head++,papantoniou2022neural} using a long video of the particular actor to reenact. Despite achieving impressive visual results, they do not generalize well to unseen identities. In this work, we mainly focus on the one-shot talking head synthesis setting, which can be applied to a new person using a single image.

\vspace*{-0.01in}
\section{Approach}
\vspace*{-0.05in}
\label{sec:approach}

\begin{figure*}[t]
    \vspace{-8pt}
    \centering
    \begin{subfigure}[t]{0.5\textwidth}
        \centering
        \includegraphics[width=\textwidth]{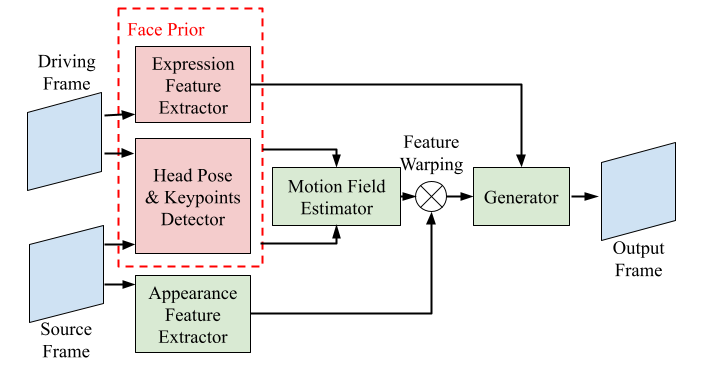}
        \caption{Proposed method.}
        \label{fig:approach_overview}
    \end{subfigure}
    \begin{subfigure}[t]{0.45\textwidth}
        \centering
        \includegraphics[width=\textwidth]{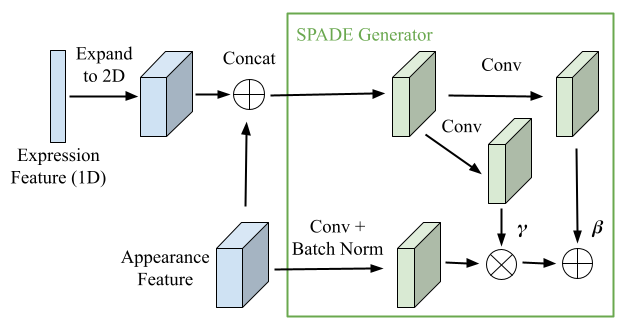}
        \caption{Generator architecture.}
        \label{fig:generator}
    \end{subfigure}
    \vspace{-4pt}
    \caption{Proposed approach. The model first extracts 1) 3D keypoints and head poses from the source and driving frame, 2) compact expression feature from the driving frame, and 3) dense appearance feature from the source frame. It then estimates the scene flows from the source to the driving frame and warps the appearance feature accordingly. Finally, the model combines the warped appearance features and the expression features to generate the output. A SPADE generator is used to combine the appearance and expression features.}
    \label{fig:approach}
    \vspace{-12pt}
\end{figure*}

We first provide an overview of the state-of-the-art Face-vid2vid model~\cite{wang2021one} for neural talking-head and the 3DMM and DECA~\cite{Feng:SIGGRAPH:2021} for face prior. Next, we introduce the proposed approach for exploiting face prior in the neural talking-head model. Finally, we describe how to perform expression manipulation in our method and the training procedure.

\subsection{Preliminary}

\noindent \textbf{Keypoint-based neural talking-head}
Our method is based on the state-of-the-art keypoint-based neural talking-head model, Face-vid2vid~\cite{wang2021one}. The Face-vid2vid model consists of four main components: appearance feature extractor, keypoint and head pose detector, motion field estimator, and image generator. Given the source and driving frames, the model first extracts the appearance feature from the source frame using the appearance feature extractor. It extracts a 3D feature volume instead of a 2D feature map to better model the 3D head motion. The model also extracts the coarse geometry from both frames using the head pose and keypoint detector, where the geometry information is encoded in sparse 3D keypoints.

Given the appearance information of the source frame and the geometry information of both frames, the model transfers the geometry from driving to the source frame using scene flow. The motion field estimator takes the appearance feature and keypoints as input and predicts the scene flow, which is then used to warp the appearance feature to match the driving frame geometry. Specifically, it first computes the flows induced by each keypoint using the zeroth order approximation in FOMM~\cite{siarohin2019first}. Because the flows induced by a keypoint are only reliable around the keypoint, the motion field estimator predicts a 3D mask indicating which flow field is more reliable in each 3D location based on the flow fields and the appearance feature. The final scene flow is obtained by combining the flow fields using the predicted mask. Finally, the generator takes the warped feature as input and predicts the target output.

During training, the model takes the source and driving frames from the same identity as inputs and uses the driving frame as the target output. The model is trained to reconstruct the target output, and the four components in the model are trained end-to-end using both image generation losses and regularization losses for the keypoint and head pose prediction. Note that the keypoint detector is trained in an unsupervised manner, and the detected keypoints do not have to correspond to conventional facial keypoints.

\noindent \textbf{Face prior from 3D face model}
Face has long been an important topic in both computer vision and graphics, and many existing approaches and models are designed for face understanding and modeling. In this work, we exploit the state-of-the-art 3D face reconstruction method, DECA~\cite{Feng:SIGGRAPH:2021}, to harness prior knowledge about 3D faces. Given a 2D face image, DECA uses an encoder to predict the identity-specific shape parameters $a \in \mathbb{R}^{100}$, identity-agnostic expression parameters $e \in \mathbb{R}^{50}$, and the head pose parameter $p \in \mathbb{R}^6$ for the FLAME model~\cite{li2017learning}. The FLAME model is then used to create a 3D mesh of the face, which provides the 3D geometry information for the 2D face. Note that 3D landmarks can be obtained from the 3D mesh directly as described in the original FLAME paper. The DECA and FLAME models allow us to extract 3D keypoints from 2D images accurately by exploiting the 3D face prior and massive 3D face data during training. Furthermore, they disentangle the identity-specific appearance and identity-agnostic expression information, which allows us to transfer generic expression across different identities.

\subsection{Neural talking-head via face prior}

This section describes how we improve the keypoint-based neural talking-head model using face prior. In particular, we replace the unsupervised keypoint detector with a supervised detector and improve the image generator by taking additional expression features from the driving frame. Both the supervised keypoint detector and the expression features are provided by the face prior model. See Fig.~\ref{fig:approach_overview}.

\noindent \textbf{Supervised keypoint detector} Instead of using unsupervised keypoints, we propose to use supervised keypoints provided by the face prior models. While unsupervised keypoints lead to a more generic and flexible approach, our empirical analysis shows that it hinders the output image quality. We observe two major problems in the unsupervised keypoints. First, the keypoints often fail to establish accurate correspondence between the source and the driving frame. Because reconstruction loss only provides indirect supervision for the keypoint detector and many unsupervised keypoints do not fall on semantically meaningful locations, it may be difficult to detect them reliably across different images. This is most common when the head poses are significantly different in the source and driving frames. The inaccurate correspondence leads to incorrect feature warping, which will degrade the reconstructed image quality and induce large geometric distortion. Second, the keypoint distribution fails to capture the importance of different face regions. In the neural talking-head problem, some face regions are more important than others in terms of perceptual quality, e.g.~mouth and eyes are more important than ears. However, unsupervised keypoint detector fails to learn this importance and distribute the keypoints evenly around the head. As a result, there are no sufficient keypoints in the important regions, and they fail to capture fine-grained motion like mouth movement. These observations suggest that supervised keypoints may improve the perceptual quality of reconstructed faces.

We extract the supervised keypoints using the face prior model as follows. Given the source $y_s$ and driving $y_d$ frames, we first extract the FLAME parameters for both images, i.e.~$[a_s, p_s, e_s]$ and $[a_d, p_d, e_d]$. We then reconstruct the face mesh using the FLAME model and extract the 3D facial keypoints. Because the FLAME parameters contain identity-specific information in the appearance parameters, it may leak the person's appearance from the driving frame to the output. To avoid this, we replace the appearance parameters of the driving frame with that of the source frame when extracting the driving frame keypoints. This approach helps to ensure that the appearance information in the driving frame cannot be passed to the output. While the original FLAME model defines 68 3D landmarks, our empirical results suggest that it is not necessary to use all of them as keypoints. Instead, we manually select 16 landmarks as our 3D keypoints if not mentioned specifically. Please refer to the supplementary material for details.

\vspace*{-0.1in}
\noindent \textbf{Expression aware generator}
While the warped features reflect the head pose and coarse expression changes, they may not be suitable for capturing fine-grained facial expressions due to the following reasons. First, the resolution of the feature is limited by the size of the feature volume, which by nature is not as scalable as 2D feature maps. Second, the sparse keypoints may not be sufficient to capture all motion in the 3D space. To address this problem, we propose to exploit the expression features obtained from the face prior model and the capacity of the generator. In practice, we feed the expression features to the generator as an additional input, which helps to capture fine-grained facial expressions that may be hard to be captured in the warped feature.

To model the interaction between the warped feature volume and the expression feature, we design the generator architecture as follows. We first expand the 1D expression features in spatial dimensions to match the shape of the warped feature map. We then concatenate the expression feature and the warped feature map as an intermediate conditional feature. Finally, a SPADE generator~\cite{park2019semantic} is used to fuse the conditional feature and the warped appearance feature to generate the outputs. See Fig.~\ref{fig:generator}. In practice, we stack eight SPADE for the generator. Also, we take the expression parameters and the jaws parameters (in the pose parameter $p_d$) extracted by the face prior model as the expression feature. Besides the head pose and keypoint detector and image generator, we follow the implementation of Face-vid2vid.

\subsection{Facial expression manipulation}
\label{sub:expression_manipulation}

While the expression feature was introduced to capture and improve fine-grained facial expression transfer, it also enables us to manipulate the facial expression semantically. Semantic editing is difficult in the original keypoint-based model, because the expression is purely controlled by the keypoints which do not correspond to semantically meaningful locations on face. In contrast, the expression features provide more direct control of facial expression.

\begin{figure}[t]
    \centering
    \includegraphics[width=\linewidth]{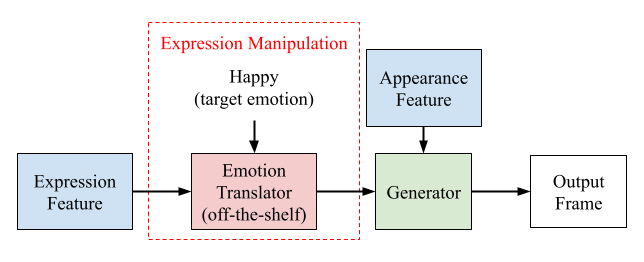}
    \vspace{-18pt}
    \caption{The proposed approach enables expression manipulation by editing the expression features. For example, we can use an off-the-shelf emotion translator to change the emotion in the output. Note that we only need to plug-in the expression manipulation module during inference, and the model does not need to be trained for this task.}
    \label{fig:emotion_translation}
    \vspace{-12pt}
\end{figure}

To achieve expression editing, we take off-the-shelf models that can alter the FLAME parameters semantically, i.e.~edit the emotion in $e$~\cite{papantoniou2022neural}. We apply the model to the expression parameters of the driving frame before feeding it to the generator. See Fig.~\ref{fig:emotion_translation}. This allows us to edit the emotion of the output image. Note that this is performed only during inference, and the talking-head model doesn't need to be trained for editing in any form. Empirical results show that this plug-and-play approach works well, and we can manipulate the emotion without affecting other facial expressions such as lip movement.

\subsection{Training}

Similar to prior works on neural talking-head~\cite{siarohin2019first,wang2021one}, we train the model using video frames sampled from talking-head videos. Given a video of a person, we randomly sample two frames as the source and driving frames respectively. Because the video contains a single person, the driving frame also serves as the target output. The appearance feature extractor, motion field estimator, and image generator are trained end-to-end using the following loss:
\begin{equation}
\mathcal{L} = \mathcal{L}_P + \mathcal{L}_G,
\end{equation}
where $\mathcal{L}_P$ is the \textbf{perceptual loss}~\cite{johnson2016perceptual} between the output and the driving frames and $\mathcal{L}_G$ is the \textbf{adversarial loss} on the output frame.  A multi-resolution patch discriminator is used for the adversarial loss. Note that unlike Face-vid2vid, we use off-the-shelf models for head pose and keypoint detection. Therefore, we do not need additional regularization losses for the detectors. We've also explored training the keypoint detector end-to-end, but we do not observe any improvement. 

We also apply random dropout to the expression feature during training to make the model robust to expression feature size. Specifically, we randomly sample $k{\in}[0, 50]$ and drop the last $k$ dimensions in expression feature before feeding it to the generator. As a result, we may use a smaller expression feature during inference. This is particularly useful in video compression applications, where we can control the bitrate by reducing the size of  the expression feature. This is similar to the keypoint dropout in Face-vid2vid and makes the model more practical in compression application.

\section{Experiments}
\label{sec:exp}

We first evaluate our method both objectively and subjectively. Next, we analyze the model performance under different bandwidth constraints for video compression applications.

\begin{table*}[t]
    \small
    \centering
    \tabcolsep=0.09cm
    \vspace{-4pt}
     \resizebox{\linewidth}{!}{
    \begin{tabular}{lcccccccccccccc}
         & \multicolumn{7}{c}{VoxCeleb} & \multicolumn{7}{c}{TalkingHead-1KH}\\
         \cmidrule(r){2-8} \cmidrule(r){9-15}
         & PSNR$\uparrow$ & SSIM$\uparrow$ & FID$\downarrow$ & AKD$\downarrow$ & AKD-M$\downarrow$ & AED$\downarrow$ & AEMOD$\downarrow$ & PSNR$\uparrow$ & SSIM$\uparrow$ & FID$\downarrow$ & AKD$\downarrow$ & AKD-M$\downarrow$ & AED$\downarrow$ & AEMOD$\downarrow$\\
         \midrule
         FOMM~\cite{siarohin2019first} & 30.4 & 0.744 & 10.7 & 1.38 & 1.32 & 0.139 & 9.31 & 30.1 & 0.607 & 21.9 & 1.85 & 1.86 & 0.162 & 6.85 \\
         Face-vid2vid~\cite{wang2021one} & \textbf{30.7} & 0.745 & 8.7 & 1.40 & 1.42 & 0.138 & 9.13 & \textbf{30.3} & \textbf{0.629} & 15.3 & 1.62 & 1.61 & 0.140 & 6.09\\
        LIA~\cite{wanglatent} & 30.6& 0.746 & 11.9 & 1.46 & 1.41 & 0.139 & 8.42  & \textbf{30.3} & 0.614 & 24.9 & 2.09 & 2.01 & 0.148 & 6.60 \\
    PIRender~\cite{ren2021pirenderer} & 29.9& 0.653 & 16.8 & 2.18 & 2.14 & 0.229 & 14.75  & 30.1 & 0.598 & 26.4 &2.11 &1.93& 0.221 & 8.97\\
         \midrule
         Ours w/o $e_{d}$ & 30.6 & 0.747 & \textbf{8.59} & 1.32 & 1.23 & 0.134 & 8.82 & 30.0 & \textbf{0.629} & 15.2 & 1.36 & 1.28 & 0.133 & 5.82\\
         Ours & \textbf{30.7} & \textbf{0.750} & 8.84 & \textbf{1.28} & \textbf{1.20} & \textbf{0.131} & \textbf{7.76} & 30.1 & 0.628 & \textbf{14.9} & \textbf{1.33} & \textbf{1.22} & \textbf{0.129} & \textbf{5.40}\\
         \bottomrule
    \end{tabular}
    }
    \vspace{-8pt}
    \caption{Quantitative comparison with state-of-the-art methods on face reconstruction.}
    \label{tab:recon_full}
    \vspace{-15pt}
\end{table*}

\noindent \textbf{Dataset} We evaluate our method on both the VoxCeleb~\cite{Nagrani17} and TalkingHead-1KH~\cite{wang2021one} dataset. We use the original train/test split of the datasets and cropped the faces from the videos and resize them to $256{\times}256$ following~\cite{siarohin2019first}.

To evaluate the model performance under large head pose difference between source and driving frames, we further collect a subset of test videos with dramatic head pose changes from VoxCeleb. Specifically, we compute the head pose variance within each video. We then take the videos with top $10\%$ variance as the hard subset to measure model performance under extreme head poses. Please refer to Appendix for more details.

\noindent \textbf{Implementation details} Our method uses a pre-extracted 3D prior to avoid repeated regeneration during training. Our method's inference time is comparable to existing baseline methods. Additional details, such as the \textbf{landmark selection process} and \textbf{computational cost}, are available in Appendix.

\noindent \textbf{Baselines} For same-identity reconstruction, we compare with FOMM~\cite{siarohin2019first} and Face-vid2vid~\cite{wang2021one}. We take the public implementation and re-train the model. Additionally, we compare with two recent direct synthesis methods (LIA~\cite{wanglatent} and PIRender~\cite{ren2021pirenderer}). For cross-identity re-enactment, we mainly compare with the keypoint-based methods (FOMM~\cite{siarohin2019first} and Face-vid2vid~\cite{wang2021one}) that use the relative motion~\cite{siarohin2019first}. That is, we transfer the motion difference between neighboring frames instead of the absolution motion. Please see Appendix for implementation details.

\noindent \textbf{Evaluation metrics} Following prior works on neural talking-head, we evaluate our method using 1) Peak signal-to-noise ratio (PSNR), Structural Similarity Index Measure (SSIM) for reconstruction faithfulness, 2) Frechet Inception Distance~\cite{heusel2017gan} (FID) for output image quality, and 3) Average Keypoint Distance~\cite{siarohin2019first} (AKD) for semantic consistency and Average Euclidean Distance~\cite{siarohin2019animating} (AED) for facial identity preserving, where AKD and AED measure the facial keypoints and facial identity by computing the similarity of keypoints and identity features extracted by off-the-shelf models. We also introduce two additional metrics: Average Keypoint Distance on Mouth (AKD-M) that measures AKD for keypoints around the mouth, and Average Emotion Distance (AEMOD) that measures the embedding distance of an off-the-shelf emotion recognition model. These two metrics are designed to capture fine-grained facial expressions. See supplementary material for details.

\begin{figure}[t]
    \vspace{-4pt}
    \centering
    \includegraphics[width=0.49\linewidth]{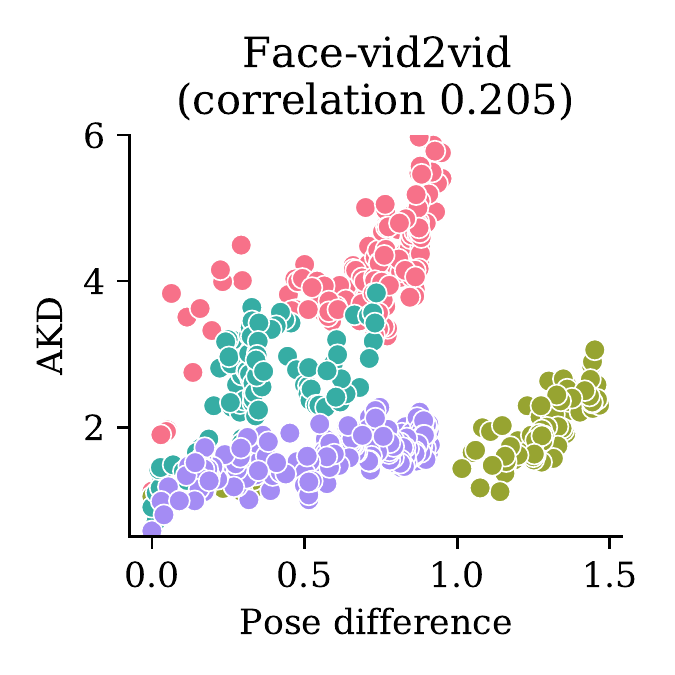}
    \includegraphics[width=0.49\linewidth]{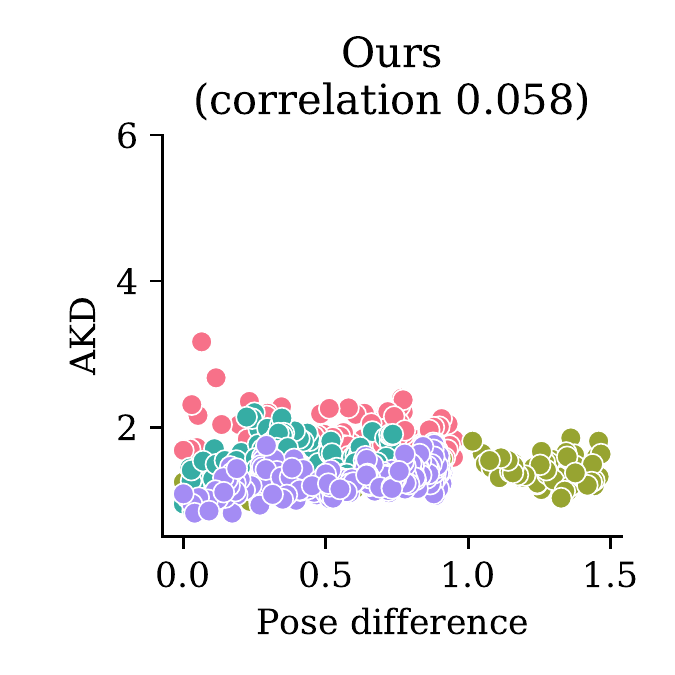}
    \vspace{-9pt}
    \caption{We analyze the correlation between reconstruction accuracy and the head pose difference between the source and driving frame. Our method has a lower correlation, indicating that it is more robust to the head pose difference. Each point corresponds to a frame, and the color indicates the video.}
    \label{fig:pair_com}
    \vspace{-16pt}
\end{figure}

\vspace*{-0.01in}
\subsection{Objective evaluation}
\vspace*{-0.01in}

\noindent \textbf{Same-identity reconstruction} We first evaluate the face reconstruction task, i.e.~the source and driving frames are from the same person. The results are in Table~\ref{tab:recon_full}. Our method consistently outperforms the baselines, particularly on semantic consistency metrics. Furthermore, the advantage is more significant on metrics that capture fine-grained facial expressions, i.e.~AKD-M and AEMOD. The results verify that our method can reconstruct fine-grained motion more faithfully, which may not be reflected in general reconstruction metrics but is crucial for perceptual quality. Also, the results show that the expression feature can consistently improve performance. The benefit is most significant in AEMOD, which suggests that the expression feature helps to transfer the emotion from the driving frame to the output. Please refer to the appendix for hard subset results. It also shows that direct synthesis methods such as LIA~\cite{wanglatent} and PIRender~\cite{ren2021pirenderer} have weaker semantic correspondence compared to our method.

To verify that our method improves the performance when there's a significant head pose difference between the source and driving frames, we further analyze the results on the hard subset. More specifically, we measure the correlation between the reconstruction accuracy and the head pose difference. The lower the correlation, the more robust the model is to head pose difference. The results are in Fig.~\ref{fig:pair_com}, where each color represents a video and each point represents a frame. The results clearly show that the accuracy of Face-vid2vid drop as the head pose difference increases. In contrast, our method is more robust, and the advantage is more significant as the pose difference increases.

\begin{table}[t]
    \small
    \centering
    \tabcolsep=0.09cm
     \resizebox{\linewidth}{!}{
    \begin{tabular}{lcccccc}
         & \multicolumn{3}{c}{VoxCeleb} & \multicolumn{3}{c}{TalkingHead-1KH}\\
         \cmidrule(r){2-4} \cmidrule(r){5-7}
         & FID$\downarrow$ & AED$\downarrow$ & AEMOD$\downarrow$ & FID$\downarrow$ & AED$\downarrow$ & AEMOD$\downarrow$\\
         \midrule
         FOMM~\cite{siarohin2019first} & 60.2 & 0.704 & 63.3 & 79.8 &0.572 & 45.2 \\
         Face-vid2vid~\cite{wang2021one} & 58.9 & 0.688 & 66.2 &  \textbf{79.7} & 0.377 & 46.3\\
         \midrule
         Ours w/o $e_{d}$ & 62.9 & \textbf{0.607} & 66.1 & 82.7& \textbf{0.332} & 46.0  \\
         Ours & \textbf{58.1} & {0.647} & \textbf{47.2} & 81.1 & {0.349} & \textbf{43.7}\\
         \bottomrule
    \end{tabular}
    }
    \vspace{-8pt}
    \caption{Quantitative comparison with state-of-the-art methods on face reenactment.}
    \label{tab:transfer_full}
    \vspace{-18pt}
\end{table}

\noindent \textbf{Cross-identity re-enactment} Next, we compare different methods on the face re-enactment task, where the source and driving frames are from different persons. The results are in Table~\ref{tab:transfer_full}. Again, our method performs better in terms of semantic consistency. AED is computed between the output and the source frame. AEMOD is computed between the output and the driving frame. The results suggest that our approach can transfer facial expressions and emotions more accurately while does not compromise source image identity-specific information.

\begin{table}[t]
    \small
    \centering
    \tabcolsep=0.12cm
    \begin{tabular}{llccc}
         & & \multicolumn{3}{c}{Prefer}\\
         \cmidrule{3-5}
         & & Ours & Baseline & Same \\
         \midrule
         \multirow{2}{*}{Reconstruction} & Photo Realism & \textbf{73.6\%} & 11.1\% & 15.3\% \\
                        & Close to Driving & \textbf{58.3\%} & 20.8\% & 20.9\% \\
         \midrule
                       & Photo Realism & \textbf{42.9\%} & 16.1\% & 41.1\% \\
          Re-enactment & Close to Driving & \textbf{74.4\%} & 12.5\% & 13.1\%\\
                       & Close to Source & 32.7\% & \textbf{36.9\%} & 30.4\%\\
         \bottomrule
    \end{tabular}
    \vspace{-8pt}
    \caption{Subjective evaluation results.}
    \label{tab:userstudy}
    \vspace{-17pt}
\end{table}

\begin{figure*}[t]
\vspace{-4pt}
\centering
\resizebox{0.83\linewidth}{!}{
    \setlength{\fboxrule}{4pt} 
     \setlength{\fboxsep}{0cm}
    \centering
    \begin{tabular}{>{\centering\arraybackslash}m{0.16\linewidth} >{\centering\arraybackslash}m{0.16\linewidth} >{\centering\arraybackslash}m{0.16\linewidth} >{\centering\arraybackslash}m{0.16\linewidth} >{\centering\arraybackslash}m{0.16\linewidth} }
         Source &    FOMM &   Face-vid2vid &   Ours &  Driving \\
        \includegraphics[ height=1.1\linewidth]{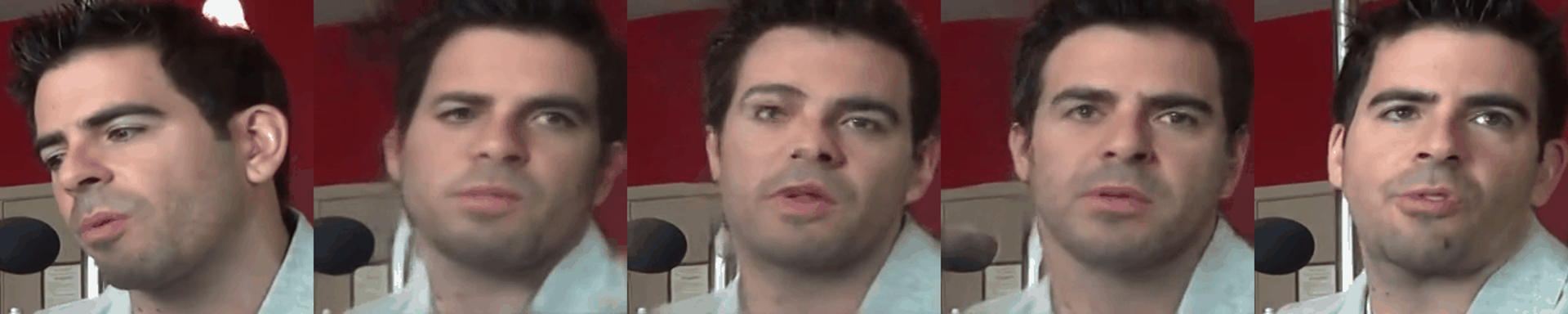}\\
        \includegraphics[ height=1.1\linewidth]{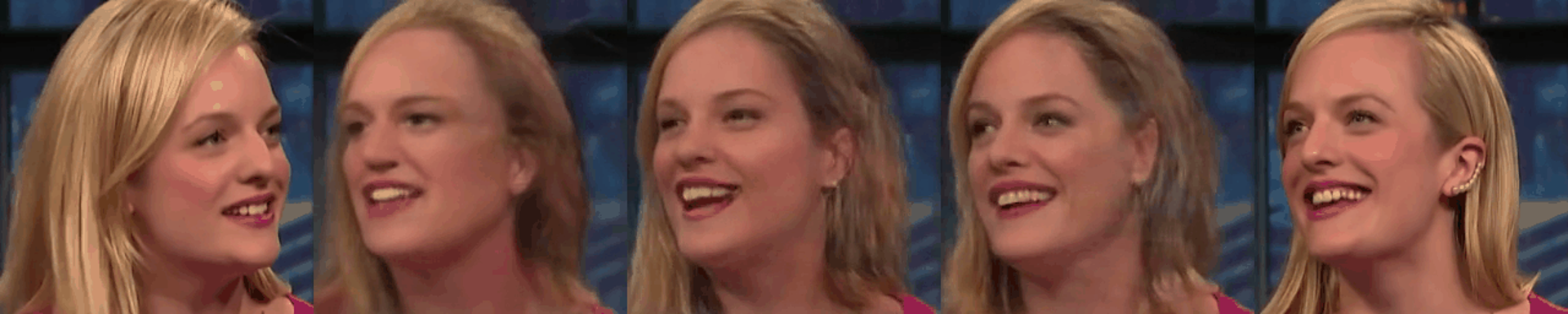}\\
        \includegraphics[ height=1.1\linewidth]{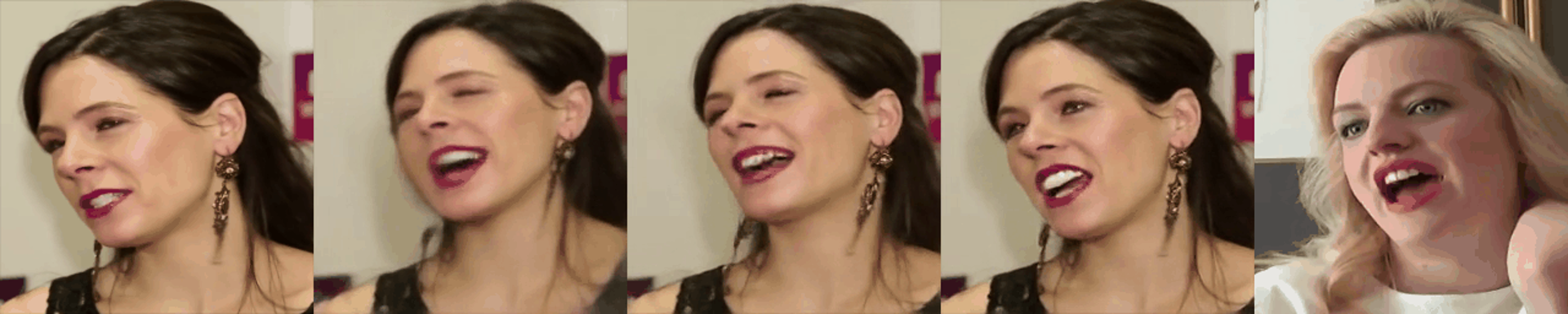}\\
        \includegraphics[ height=1.1\linewidth]{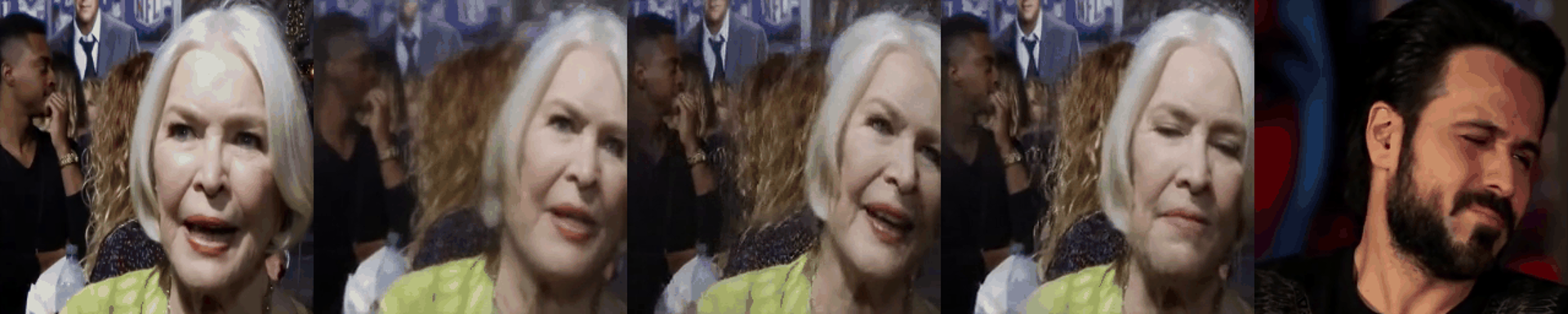}\\
        \includegraphics[ height=1.1\linewidth]{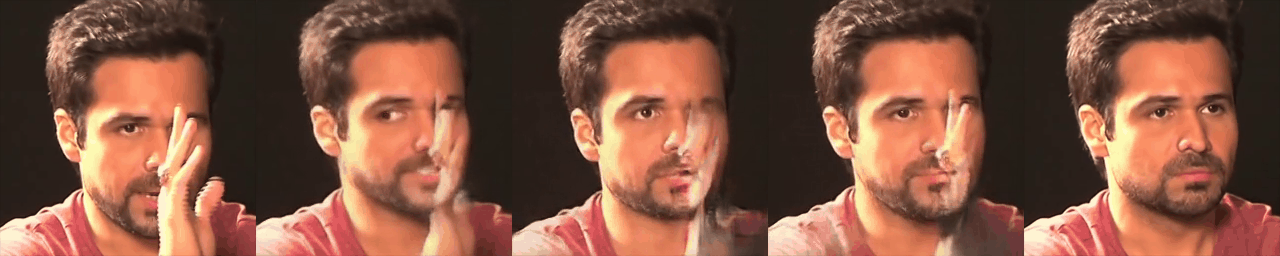}\\
     \end{tabular}  
}
\vspace{-12pt}
\caption{Qualitative examples. The first and second rows show reconstruction results, and the third and fourth rows show the re-enactment results. For a fair comparison with FOMM~\cite{siarohin2019first}, visual examples for re-enactment are reported via the relative motion adopted in FOMM. Compared with baselines, our method generates more realistic outputs especially when the head poses are significantly different in the source and driving frames. Also, it transfers fine-grained expression more faithfully. The last row shows a failure example, where our method struggles with occlusion. See supplementary material for video examples.}
\label{fig:compare_qualitative}
\vspace{-7pt} 
\end{figure*}

\subsection{Subjective evaluation} 
We further augment our quantitative evaluation with a user study. In the user study, we show the videos synthesized by two different methods together with the source and driving videos to the raters. The raters are asked to answer 1) which video is more realistic, 2) which video is more similar to the driving video in terms of the subject's pose and facial expression, and 3) which video is more similar to the source frame in terms of the subject's identity in re-enactment task. We compare our method with Face-vid2vid in the user study, and the order of the two synthesized videos are chosen randomly to avoid bias. We randomly select 25 videos for reconstruction and 20 videos for re-enactment, and each video is rated by 10 raters.

The results are in Table~\ref{tab:userstudy}. Our method is preferred by raters most of the time. The only exception is the similarity with the source frame in the re-enactment task, where the two methods are almost not distinguishable. The results further verify the superior perceptual quality of our method. Interestingly, our method outperforms the baseline with a much larger margin in the subjective evaluation. This suggests that fine-grained facial expression, while introducing only minor differences in pixel and feature space, can lead to a significant difference in perceptual quality. Therefore, it is important to capture fine-grained details in face video synthesis, which further justifies our motivation.

\begin{figure*}[t]
\vspace{-4pt}
\resizebox{0.8\linewidth}{!}{
    \setlength{\fboxrule}{4pt} 
     \setlength{\fboxsep}{0cm}
    \centering
    \begin{tabular}{>{\centering\arraybackslash}m{0.16\linewidth} >{\centering\arraybackslash}m{0.16\linewidth} >{\centering\arraybackslash}m{0.16\linewidth} >{\centering\arraybackslash}m{0.16\linewidth} >{\centering\arraybackslash}m{0.16\linewidth} }
          Source & Driving    & Unmodified& Neutral &    Happy    \\
                \includegraphics[  width=1\linewidth]{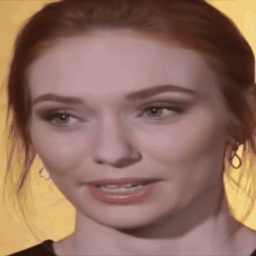} & \includegraphics[  width=1\linewidth]{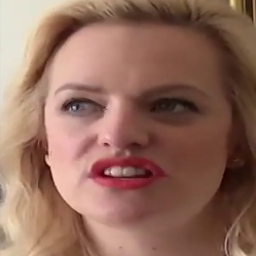} & \includegraphics[  width=1\linewidth]{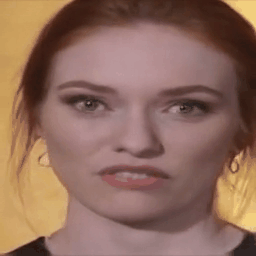} & \includegraphics[  width=1\linewidth]{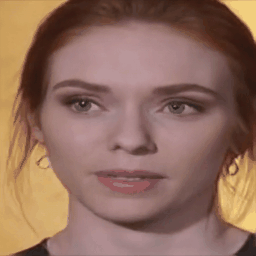} & \includegraphics[  width=1\linewidth]{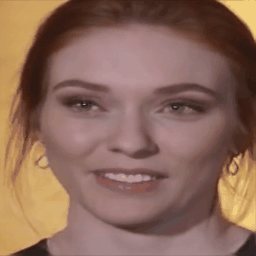}\\
                \includegraphics[  width=1\linewidth]{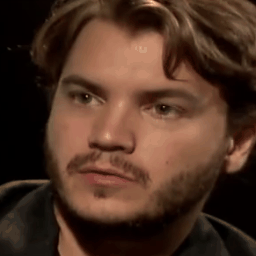} & \includegraphics[  width=1\linewidth]{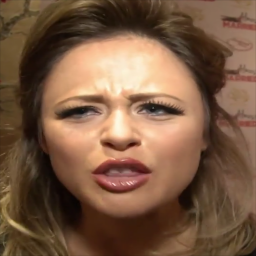} & \includegraphics[  width=1\linewidth]{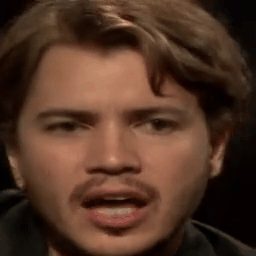} & \includegraphics[  width=1\linewidth]{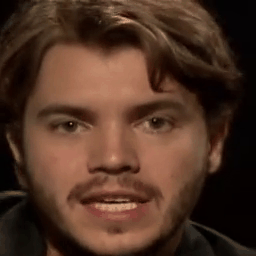} & \includegraphics[  width=1\linewidth]{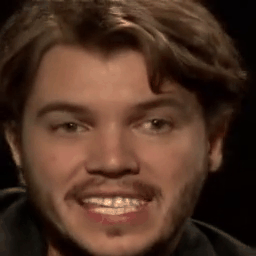}\\

     \end{tabular}  
}
\centering
\vspace{-10pt}
\caption{Facial expression manipulation results. Our method can simultaneously transfer the coarse head motion from the driving frame and alter the fine-grained expression using the expression feature. Note that the results are obtained by plugging in an off-the-shelf emotion translation model during inference without any change to the model.}
\label{fig:compare_qualitative_emo}
\vspace{-9pt}
\end{figure*}

Fig.~\ref{fig:compare_qualitative} shows the qualitative results, where we can see that our method can transfer the facial expression from the driving frame to the output faithfully. We also show the result of emotion manipulation in Fig.~\ref{fig:compare_qualitative_emo}. As motioned in Sec.~\ref{sub:expression_manipulation}, one benefit of the proposed method is that it enables facial expression manipulation. We verify this by changing the facial expression to edit the emotion despite the fact that the model has never been trained for this task.

Despite the promising performance, one limitation of our method is that it is not scalable w.r.t.~image resolution. We were unable to apply it to synthesize a high resolution (e.g.~$1024{\times}1024$) face video due to the overhead introduced by 3D feature volume. While prior works mainly focus on face video with relatively low resolution, i.e.~$256{\times}256$, there's a clear need for synthesizing high-resolution face videos in applications. Also, our model may fail when the face is occluded. See Fig.~\ref{fig:compare_qualitative}.

\begin{figure}[t]
    \centering
    \includegraphics[width=0.9\linewidth]{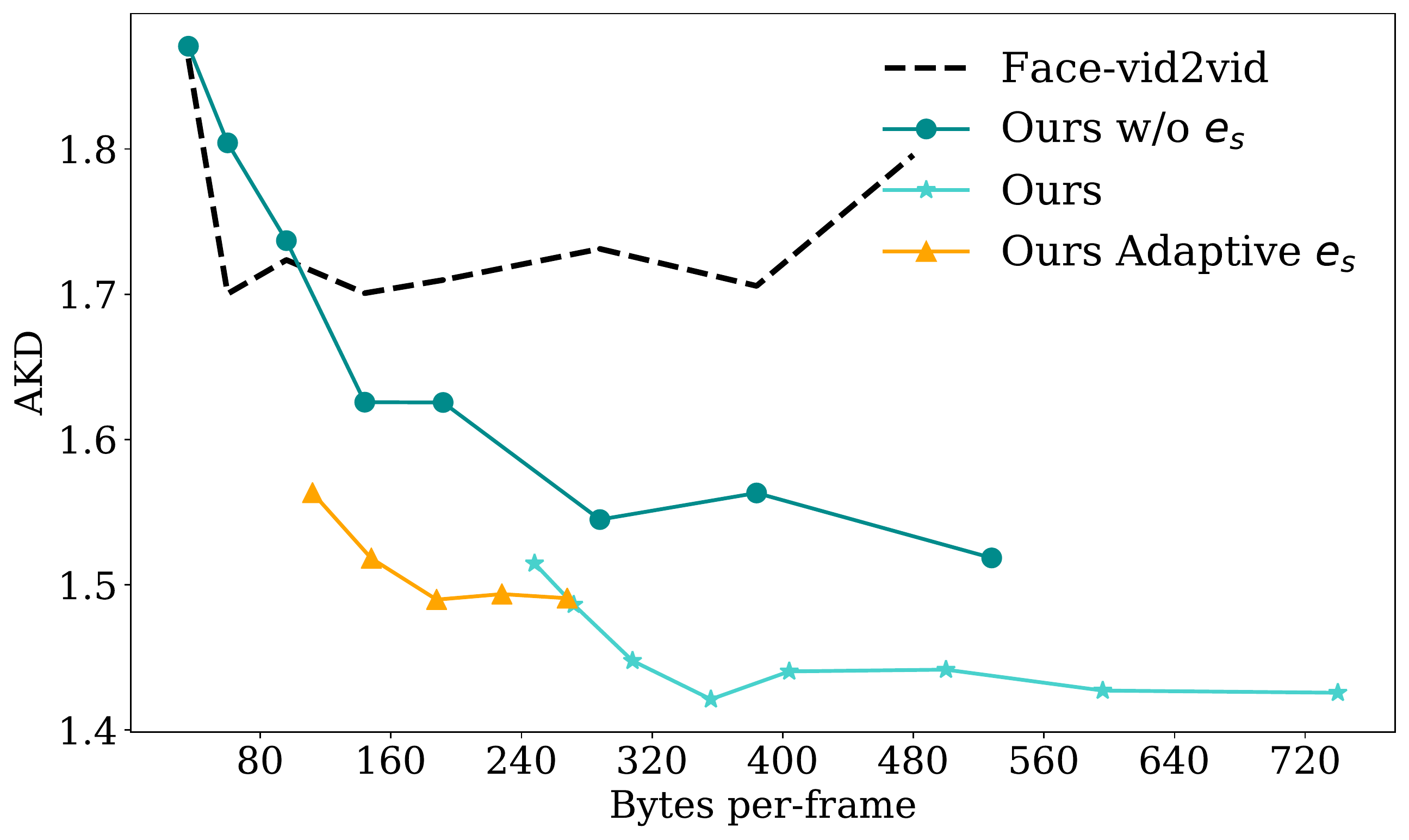}
    \vspace{-9pt}
    \caption{Reconstruction accuracy versus bitrate. The driving video can be encoded using the keypoints and expression features, and we control the bitrate by changing the number of keypoints and size of $e_d$ (for Ours adaptive $e_d$). Our method achieves the best reconstruction accuracy under all bitrates.}
    \label{fig:recon_sub}
    \vspace{-15pt}
\end{figure}

\subsection{Output quality versus bandwidth}
Finally, we analyze how the number of keypoints and the size of the expression feature affects the model accuracy. This is particularly important because video compression is an important application of neural talking-head, and one benefit of the keypoint-based method is that it allows us to adjust the bitrate by altering the number of keypoints. Note that compression is achieved by encoding the driving video into the keypoints and expression features, and the video can be reconstructed using the model. In this analysis, we compare the following methods: 1) Face-vid2vid with a variable number of keypoints, 2) Ours w/o $e_d$ with a variable number of keypoints, 3) Ours with a variable number of keypoints, and 4) Ours with variable expression feature size. We measure the model accuracy using AKD and the bitrate using the number of bytes it takes to represent each frame.

To test the difference for the more challenging videos, we shows the AKD results on the hard subset in Fig.~\ref{fig:recon_sub} (Please refers to the supplementary material for other metrics and results on the full testset). We can see that while the performance of Face-vid2vid improves as the number of keypoints increases, it saturates quickly at around 15 keypoints and will degrade if we keep increasing the number of keypoints. The result suggests that, unlike traditional compression algorithms, we cannot improves the model performance by simply increasing the number of keypoints. In contrast, Ours w/o $e_{d}$ outperforms Face-vid2vid in higher bitrate scenarios, and the performance will converge rather than degrade as the number of keypoints increases. The result suggests that the supervised keypoints lead to both a better absolute performance as well as better scalability with bitrate. Finally, the results justify the benefit of expression feature in terms of compression, where we can achieve the best reconstruction accuracy under all bitrates when we adjust the keypoint number and expression feature size jointly.

\vspace*{-0.05in}
\section{Conclusion}
\vspace*{-0.05in}

This work studies how to improve the neural talking-head model using 3D face prior. Our analysis show the limitations of the unsupervised keypoint detector used in the current state-of-the-art model. We propose to address these limitations by using 3D face model for more accurate keypoint detection and fine-grained facial expression modeling. Empirical results show that our method consistently outperforms the state-of-the-art model in various metrics under all compression rates, and the output video is preferred by users more than $70\%$ of the time in the user study.

\newpage

{\small
\bibliographystyle{ieee_fullname}
\bibliography{egbib}

\begin{thebibliography}{10}\itemsep=-1pt

\bibitem{averbuch2017bringing}
Hadar Averbuch-Elor, Daniel Cohen-Or, Johannes Kopf, and Michael~F Cohen.
\newblock Bringing portraits to life.
\newblock {\em ACM Transactions on Graphics (ToG)}, 36(6):1--13, 2017.

\bibitem{blanz1999morphable}
Volker Blanz and Thomas Vetter.
\newblock A morphable model for the synthesis of 3d faces.
\newblock In {\em Proceedings of the 26th annual conference on Computer
  graphics and interactive techniques}, pages 187--194, 1999.

\bibitem{bulat2017far}
Adrian Bulat and Georgios Tzimiropoulos.
\newblock How far are we from solving the 2d \& 3d face alignment problem? (and
  a dataset of 230,000 3d facial landmarks).
\newblock In {\em International Conference on Computer Vision}, 2017.

\bibitem{burkov2020neural}
Egor Burkov, Igor Pasechnik, Artur Grigorev, and Victor Lempitsky.
\newblock Neural head reenactment with latent pose descriptors.
\newblock In {\em Proceedings of the IEEE/CVF conference on computer vision and
  pattern recognition}, pages 13786--13795, 2020.

\bibitem{deng2008computer}
Zhigang Deng and Junyong Noh.
\newblock Computer facial animation: A survey.
\newblock In {\em Data-driven 3D facial animation}, pages 1--28. Springer,
  2008.

\bibitem{doukas2021head2head++}
Michail~Christos Doukas, Mohammad~Rami Koujan, Viktoriia Sharmanska, Anastasios
  Roussos, and Stefanos Zafeiriou.
\newblock Head2head++: Deep facial attributes re-targeting.
\newblock {\em IEEE Transactions on Biometrics, Behavior, and Identity
  Science}, 3(1):31--43, 2021.

\bibitem{doukas2021headgan}
Michail~Christos Doukas, Stefanos Zafeiriou, and Viktoriia Sharmanska.
\newblock Headgan: One-shot neural head synthesis and editing.
\newblock In {\em Proceedings of the IEEE/CVF International Conference on
  Computer Vision}, pages 14398--14407, 2021.

\bibitem{esser2018variational}
Patrick Esser, Ekaterina Sutter, and Bj{\"o}rn Ommer.
\newblock A variational u-net for conditional appearance and shape generation.
\newblock In {\em Proceedings of the IEEE conference on computer vision and
  pattern recognition}, pages 8857--8866, 2018.

\bibitem{Feng:SIGGRAPH:2021}
Yao Feng, Haiwen Feng, Michael~J. Black, and Timo Bolkart.
\newblock Learning an animatable detailed {3D} face model from in-the-wild
  images.
\newblock {\em ACM Transactions on Graphics (ToG), Proc. SIGGRAPH},
  40(4):88:1--88:13, Aug. 2021.

\bibitem{geng2018warp}
Jiahao Geng, Tianjia Shao, Youyi Zheng, Yanlin Weng, and Kun Zhou.
\newblock Warp-guided gans for single-photo facial animation.
\newblock {\em ACM Transactions on Graphics (ToG)}, 37(6):1--12, 2018.

\bibitem{ha2020marionette}
Sungjoo Ha, Martin Kersner, Beomsu Kim, Seokjun Seo, and Dongyoung Kim.
\newblock Marionette: Few-shot face reenactment preserving identity of unseen
  targets.
\newblock In {\em Proceedings of the AAAI conference on artificial
  intelligence}, volume~34, pages 10893--10900, 2020.

\bibitem{heusel2017gan}
Martin Heusel, Hubert Ramsauer, Thomas Unterthiner, Bernhard Nessler, and Sepp
  Hochreiter.
\newblock Gans trained by a two time-scale update rule converge to a local nash
  equilibrium.
\newblock In I. Guyon, U.~Von Luxburg, S. Bengio, H. Wallach, R. Fergus, S.
  Vishwanathan, and R. Garnett, editors, {\em Advances in Neural Information
  Processing Systems}, volume~30. Curran Associates, Inc., 2017.

\bibitem{heusel2017gans}
Martin Heusel, Hubert Ramsauer, Thomas Unterthiner, Bernhard Nessler, and Sepp
  Hochreiter.
\newblock Gans trained by a two time-scale update rule converge to a local nash
  equilibrium.
\newblock {\em Advances in neural information processing systems}, 30, 2017.

\bibitem{hong2022depth}
Fa-Ting Hong, Longhao Zhang, Li Shen, and Dan Xu.
\newblock Depth-aware generative adversarial network for talking head video
  generation.
\newblock In {\em Proceedings of the IEEE/CVF Conference on Computer Vision and
  Pattern Recognition}, pages 3397--3406, 2022.

\bibitem{johnson2016perceptual}
Justin Johnson, Alexandre Alahi, and Li Fei-Fei.
\newblock Perceptual losses for real-time style transfer and super-resolution.
\newblock In {\em European conference on computer vision}, pages 694--711.
  Springer, 2016.

\bibitem{kim2018deep}
Hyeongwoo Kim, Pablo Garrido, Ayush Tewari, Weipeng Xu, Justus Thies, Matthias
  Niessner, Patrick P{\'e}rez, Christian Richardt, Michael Zollh{\"o}fer, and
  Christian Theobalt.
\newblock Deep video portraits.
\newblock {\em ACM Transactions on Graphics (TOG)}, 37(4):1--14, 2018.

\bibitem{koujan2020head2head}
Mohammad~Rami Koujan, Michail~Christos Doukas, Anastasios Roussos, and Stefanos
  Zafeiriou.
\newblock Head2head: Video-based neural head synthesis.
\newblock In {\em 2020 15th IEEE International Conference on Automatic Face and
  Gesture Recognition (FG 2020)}, pages 16--23. IEEE, 2020.

\bibitem{li2017learning}
Tianye Li, Timo Bolkart, Michael~J Black, Hao Li, and Javier Romero.
\newblock Learning a model of facial shape and expression from 4d scans.
\newblock {\em ACM Trans. Graph.}, 36(6):194--1, 2017.

\bibitem{liang2022expressive}
Borong Liang, Yan Pan, Zhizhi Guo, Hang Zhou, Zhibin Hong, Xiaoguang Han, Junyu
  Han, Jingtuo Liu, Errui Ding, and Jingdong Wang.
\newblock Expressive talking head generation with granular audio-visual
  control.
\newblock In {\em Proceedings of the IEEE/CVF Conference on Computer Vision and
  Pattern Recognition}, pages 3387--3396, 2022.

\bibitem{mallya2022implicit}
Arun Mallya, Ting-Chun Wang, and Ming-Yu Liu.
\newblock Implicit warping for animation with image sets.
\newblock {\em arXiv preprint arXiv:2210.01794}, 2022.

\bibitem{meshry2021learned}
Moustafa Meshry, Saksham Suri, Larry~S Davis, and Abhinav Shrivastava.
\newblock Learned spatial representations for few-shot talking-head synthesis.
\newblock In {\em Proceedings of the IEEE/CVF International Conference on
  Computer Vision}, pages 13829--13838, 2021.

\bibitem{Nagrani17}
A. Nagrani, J.~S. Chung, and A. Zisserman.
\newblock Voxceleb: a large-scale speaker identification dataset.
\newblock In {\em INTERSPEECH}, 2017.

\bibitem{oquab2021low}
Maxime Oquab, Pierre Stock, Daniel Haziza, Tao Xu, Peizhao Zhang, Onur Celebi,
  Yana Hasson, Patrick Labatut, Bobo Bose-Kolanu, Thibault Peyronel, et~al.
\newblock Low bandwidth video-chat compression using deep generative models.
\newblock In {\em Proceedings of the IEEE/CVF Conference on Computer Vision and
  Pattern Recognition}, pages 2388--2397, 2021.

\bibitem{papantoniou2022neural}
Foivos~Paraperas Papantoniou, Panagiotis~P Filntisis, Petros Maragos, and
  Anastasios Roussos.
\newblock Neural emotion director: Speech-preserving semantic control of facial
  expressions in" in-the-wild" videos.
\newblock In {\em Proceedings of the IEEE/CVF Conference on Computer Vision and
  Pattern Recognition}, pages 18781--18790, 2022.

\bibitem{park2019semantic}
Taesung Park, Ming-Yu Liu, Ting-Chun Wang, and Jun-Yan Zhu.
\newblock Semantic image synthesis with spatially-adaptive normalization.
\newblock In {\em Proceedings of the IEEE/CVF conference on computer vision and
  pattern recognition}, pages 2337--2346, 2019.

\bibitem{ren2021pirenderer}
Yurui Ren, Ge Li, Yuanqi Chen, Thomas~H Li, and Shan Liu.
\newblock Pirenderer: Controllable portrait image generation via semantic
  neural rendering.
\newblock In {\em Proceedings of the IEEE/CVF International Conference on
  Computer Vision}, pages 13759--13768, 2021.

\bibitem{Savchenko_2022_CVPRW}
Andrey~V. Savchenko.
\newblock Video-based frame-level facial analysis of affective behavior on
  mobile devices using efficientnets.
\newblock In {\em Proceedings of the IEEE/CVF Conference on Computer Vision and
  Pattern Recognition (CVPR) Workshops}, pages 2359--2366, June 2022.

\bibitem{sha2021deep}
Tong Sha, Wei Zhang, Tong Shen, Zhoujun Li, and Tao Mei.
\newblock Deep person generation: A survey from the perspective of face, pose
  and cloth synthesis.
\newblock {\em arXiv preprint arXiv:2109.02081}, 2021.

\bibitem{shu2022few}
Changyong Shu, Hemao Wu, Hang Zhou, Jiaming Liu, Zhibin Hong, Changxing Ding,
  Junyu Han, Jingtuo Liu, Errui Ding, and Jingdong Wang.
\newblock Few-shot head swapping in the wild.
\newblock In {\em Proceedings of the IEEE/CVF Conference on Computer Vision and
  Pattern Recognition}, pages 10789--10798, 2022.

\bibitem{siarohin2019animating}
Aliaksandr Siarohin, St{\'e}phane Lathuili{\`e}re, Sergey Tulyakov, Elisa
  Ricci, and Nicu Sebe.
\newblock Animating arbitrary objects via deep motion transfer.
\newblock In {\em Proceedings of the IEEE/CVF Conference on Computer Vision and
  Pattern Recognition}, pages 2377--2386, 2019.

\bibitem{siarohin2019first}
Aliaksandr Siarohin, St{\'e}phane Lathuili{\`e}re, Sergey Tulyakov, Elisa
  Ricci, and Nicu Sebe.
\newblock First order motion model for image animation.
\newblock {\em Advances in Neural Information Processing Systems}, 32, 2019.

\bibitem{siarohin2021motion}
Aliaksandr Siarohin, Oliver~J Woodford, Jian Ren, Menglei Chai, and Sergey
  Tulyakov.
\newblock Motion representations for articulated animation.
\newblock In {\em Proceedings of the IEEE/CVF Conference on Computer Vision and
  Pattern Recognition}, pages 13653--13662, 2021.

\bibitem{tao2022structure}
Jiale Tao, Biao Wang, Borun Xu, Tiezheng Ge, Yuning Jiang, Wen Li, and Lixin
  Duan.
\newblock Structure-aware motion transfer with deformable anchor model.
\newblock In {\em Proceedings of the IEEE/CVF Conference on Computer Vision and
  Pattern Recognition}, pages 3637--3646, 2022.

\bibitem{wang2019few}
Ting-Chun Wang, Ming-Yu Liu, Andrew Tao, Guilin Liu, Jan Kautz, and Bryan
  Catanzaro.
\newblock Few-shot video-to-video synthesis.
\newblock {\em arXiv preprint arXiv:1910.12713}, 2019.

\bibitem{wang2021one}
Ting-Chun Wang, Arun Mallya, and Ming-Yu Liu.
\newblock One-shot free-view neural talking-head synthesis for video
  conferencing.
\newblock In {\em Proceedings of the IEEE/CVF conference on computer vision and
  pattern recognition}, pages 10039--10049, 2021.

\bibitem{wanglatent}
Yaohui Wang, Di Yang, Francois Bremond, and Antitza Dantcheva.
\newblock Latent image animator: Learning to animate images via latent space
  navigation.
\newblock In {\em International Conference on Learning Representations}.

\bibitem{xu2022designing}
Chao Xu, Jiangning Zhang, Yue Han, Guanzhong Tian, Xianfang Zeng, Ying Tai,
  Yabiao Wang, Chengjie Wang, and Yong Liu.
\newblock Designing one unified framework for high-fidelity face reenactment
  and swapping.
\newblock In {\em European Conference on Computer Vision}, pages 54--71.
  Springer, 2022.

\bibitem{yao2020mesh}
Guangming Yao, Yi Yuan, Tianjia Shao, and Kun Zhou.
\newblock Mesh guided one-shot face reenactment using graph convolutional
  networks.
\newblock In {\em Proceedings of the 28th ACM International Conference on
  Multimedia}, pages 1773--1781, 2020.

\bibitem{zakharov2020fast}
Egor Zakharov, Aleksei Ivakhnenko, Aliaksandra Shysheya, and Victor Lempitsky.
\newblock Fast bi-layer neural synthesis of one-shot realistic head avatars.
\newblock In {\em European Conference on Computer Vision}, pages 524--540.
  Springer, 2020.

\bibitem{zakharov2019few}
Egor Zakharov, Aliaksandra Shysheya, Egor Burkov, and Victor Lempitsky.
\newblock Few-shot adversarial learning of realistic neural talking head
  models.
\newblock In {\em Proceedings of the IEEE/CVF international conference on
  computer vision}, pages 9459--9468, 2019.

\bibitem{zengfnevr}
Bohan Zeng, Boyu Liu, Hong Li, Xuhui Liu, Jianzhuang Liu, Dapeng Chen, Wei
  Peng, and Baochang Zhang.
\newblock Fnevr: Neural volume rendering for face animation.
\newblock {\em arXiv preprint arXiv:2209.10340}, 2022.

\bibitem{zhang2021flow}
Zhimeng Zhang, Lincheng Li, Yu Ding, and Changjie Fan.
\newblock Flow-guided one-shot talking face generation with a high-resolution
  audio-visual dataset.
\newblock In {\em Proceedings of the IEEE/CVF Conference on Computer Vision and
  Pattern Recognition}, pages 3661--3670, 2021.

\bibitem{zhao2022thin}
Jian Zhao and Hui Zhang.
\newblock Thin-plate spline motion model for image animation.
\newblock In {\em Proceedings of the IEEE/CVF Conference on Computer Vision and
  Pattern Recognition}, pages 3657--3666, 2022.

\bibitem{zhong2022geometry}
Yatao Zhong, Faezeh Amjadi, and Ilya Zharkov.
\newblock Geometry driven progressive warping for one-shot face animation.
\newblock {\em arXiv preprint arXiv:2210.02391}, 2022.

\end{thebibliography}
}

\newpage

\appendix

\onecolumn 
\section*{Appendix}
\label{sec:appendix}

\noindent The appendix is organized as follows:

\paragraph{Appendix~\ref{app:experiments}} we include additional descriptions of the datasets (Appendix~\ref{append:dataset}), metrics (Appendix~\ref{append:metrics})

\paragraph{Appendix~\ref{subsec:impl}} we include the implementation details: the network architectures and optimization details.

\paragraph{Appendix~\ref{app:results}} we include additional results. In \ref{app:visexample}, we provide additional qualitative examples of the model performance for reconstruction, cross-person reenactment, emotion editing, and pose editing.  In \ref{app:quanres}, we provide additional quantitative comparison results.

Additionally, we provide the video examples of reconstruction, re-enactment and emotion manipulation in the folder.

\section{Experiments}
\label{app:experiments}
\subsection{Dataset}
\label{append:dataset}

\paragraph{VoxCeleb dataset~\cite{Nagrani17}} is a face dataset of 22496 videos. It is extracted from YouTube videos, consisting of interview videos of different celebrities. For pre-processing, we follow the steps as specified in~\cite{siarohin2019first}. Specifically, we extract an initial bounding box in the first video frame and track the face until it is too far away from the initial position. Then, we crop the frames using the smallest crop containing all the bounding boxes. The process is repeated until the end of the sequence. Videos with a resolution lower than 256 $\times$ 256 are filtered out sequences. The remaining videos are resized to 256 $\times$ 256.  Overall, we obtain 12331 training videos and 522 test videos, with lengths varying from 64 to 1024 frames.

To evaluate the model performance in reconstructing the more challenging videos with large head poses, we take the videos with top $10\%$ pose variance among the full testset as the hard subset for evaluation. Specifically, we use the DECA~\cite{Feng:SIGGRAPH:2021} model to obtain the head poses for the video frames and compute the variance within each video. We then sort the videos according to such variance in a large to a small order. To this end, we select the $10\%$ videos with the largest variance out to form the hard subset for evaluation.

\paragraph{TalkingHead-1KH~\cite{wang2021one}} is a large dataset obtained via various sources, where a large portion of them is from YouTube websites. We follow the same preprocessing pipeline as~\cite{wang2021one}. Since some video links are no longer effective, we collect over 350k video clips for training. We use the evaluation set with 37 videos. All the videos are resized to 256 $\times$ 256.

\subsection{Metrics}
\label{append:metrics}

First, we quantitatively evaluate each method on the task of video reconstruction, which consists of reconstructing the input video from a representation in which appearance and motion are decoupled. Specifically, we reconstruct the input video by combining the sparse keypoints and lightweight expression code (if available) from each frame and the appearance feature extracted from the source frame. For the full testset, we follow the previous method to use the first frame within each video as the source frame. For the hard subset, we take the frame with the largest head pose within each video as the source frame.

 \paragraph{Peak signal-to-noise ratio (PSNR):} PSNR measures the image reconstruction quality by computing the mean squared error (MSE) between the ground truth and the reconstructed image.
    \paragraph{Structural Similarity Index Measure (SSIM):} SSIM measures the structural similarity between patches of the input images. Therefore, it is more robust to global illumination changes than PSNR-based absolute errors.
    \paragraph{Average keypoint distance (AKD):} The metric is used to evaluate the semantic consistency between the generated video and the ground truth video. We employ an external keypoint detector~\cite{bulat2017far} to detect the keypoints from the ground truth and generated video. We then return the average distance between the detected keypoints of the ground truth and the generated video as the AKD value.
    \paragraph{Average Keypoint Distance on Mouth (AKD-M):} The metric is used to evaluate the semantic consistency between the generated video and the ground truth video around the mouth, which is crucial for the speech-preserving ability of the generated video. We use the same external keypoint detector~\cite{bulat2017far} to detect the keypoints, where only 20 keypoints that belong to the mouth region are selected. We then return the average distance between the 20 keypoints of the ground truth and the generated video as the AKD-M value.
    \paragraph{Average Euclidean Distance (AED):} This metric is designed to reflect how well the identity is preserved. We adopt the feature-based metric employed in~\cite{esser2018variational} to extract the identity feature. It computes the Average Euclidean Distance (AED) between a feature representation of the ground truth and the generated video frames. 
    \paragraph{Frechet Inception Distance (FID): } FID measures the distance between the distributions of synthesized and real images~\cite{heusel2017gans}. It aims to evaluate the quality of individual frames.
    \paragraph{Average EMotion Distance (AEMOD): } AEMOD is designed to evaluate the emotion-preserving performance between the generated video and the ground truth video. Specifically, we use a pretrained emotion recognition network~\cite{Savchenko_2022_CVPRW} to extract feature representations from the frames of the generated video and the ground truth video.  We then compute the average euclidean distance between the feature representation of the ground truth and the generated video frames.

We also evaluate the model performance for cross-person re-enactment via FID, AED, and AEMOD. For the FID, we measure the distribution distance between the generated frames and the original frames in the full testset. For the AED computation, since there is no ground truth frame, we take the following strategy to measure the identity-preserving ability. For each generated frame, we compute the euclidean distance of the identity feature representation between the generated frame and all the frames within the same video as the source frame. We then take the average of the euclidean distance to form the AED. This aims to improve the robustness of the metric. For the AEMOD computation, we use the pretrained emotion recognition network~\cite{Savchenko_2022_CVPRW} to extract the features from the generated frame and the corresponding driving frame. We then compute the euclidean distance between each pair. To this end, we average the results for all the pairs to form the AEMOD.

\subsection{Network architectures}

\section{Implementation Details}
\label{subsec:impl}

We adopt the official code from the FOMM~\cite{siarohin2019first} and the widely used public implementation from Face-vid2vid. For evaluation of the re-enactment, we follow the relative motion~\cite{siarohin2019first} to compare with FOMM~\cite{siarohin2019first} in a fair way. Transferring relative motion over absolute coordinates allows transferring only relevant motion patterns while preserving global object geometry.

\subsection{Network architectures.} 
\begin{figure*}[t]
\vspace{-4pt}
\centering
\includegraphics[width=0.27\linewidth]{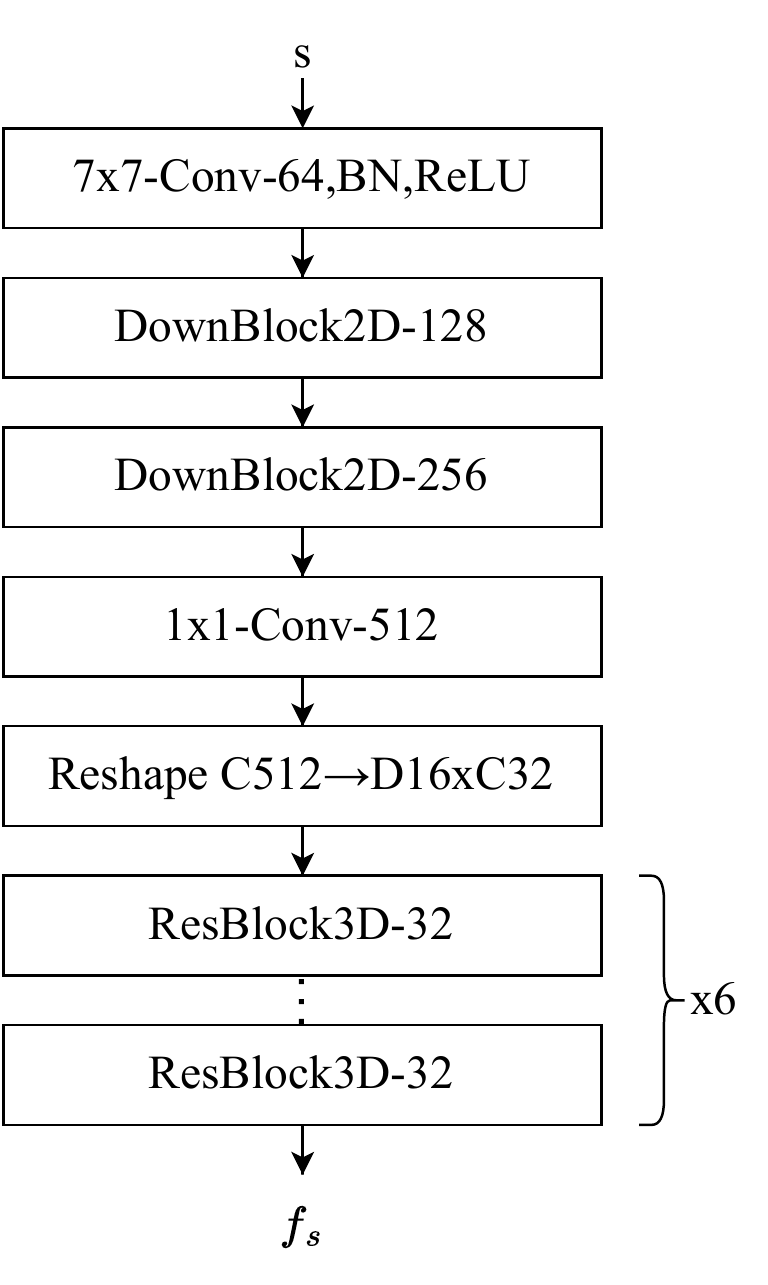}
\includegraphics[width=0.3\linewidth]{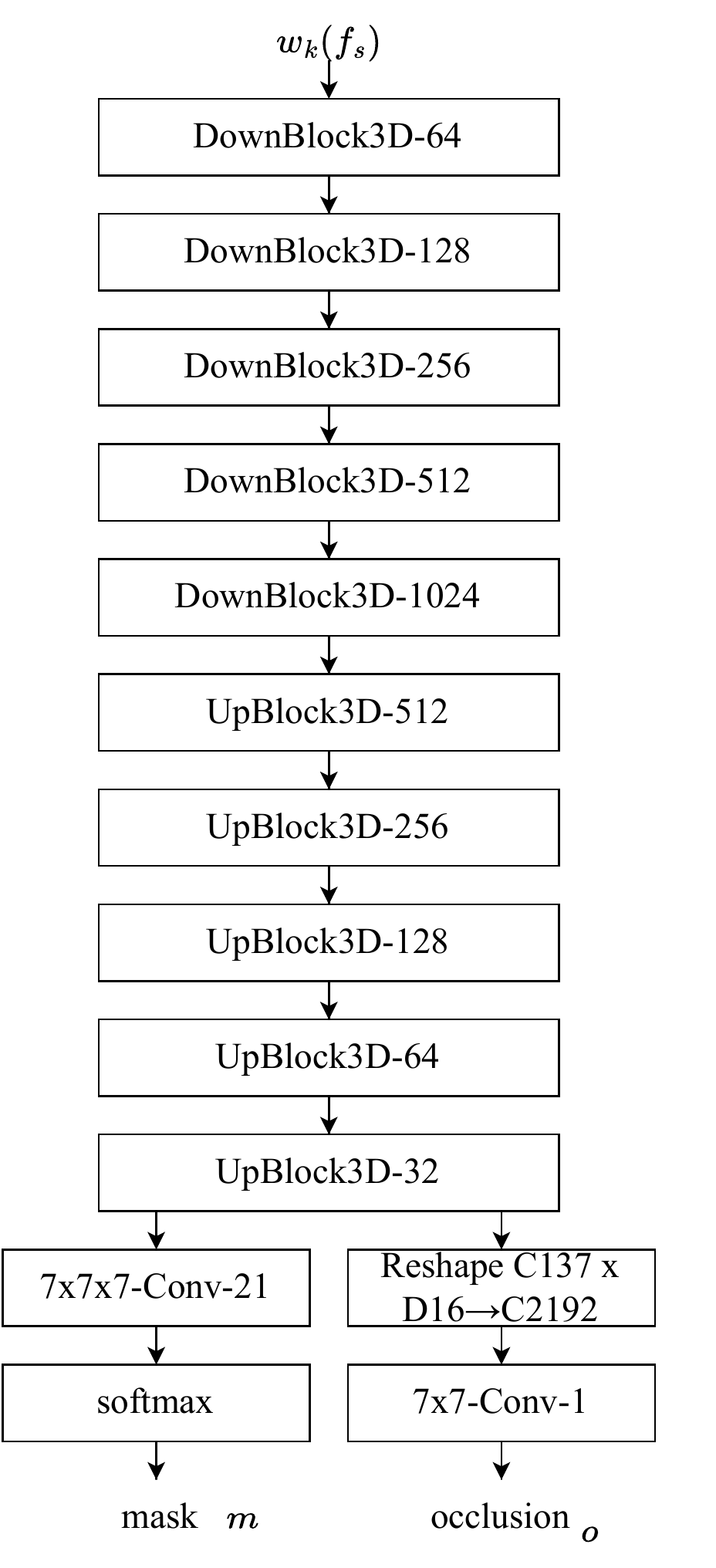}
\includegraphics[width=0.37\linewidth]{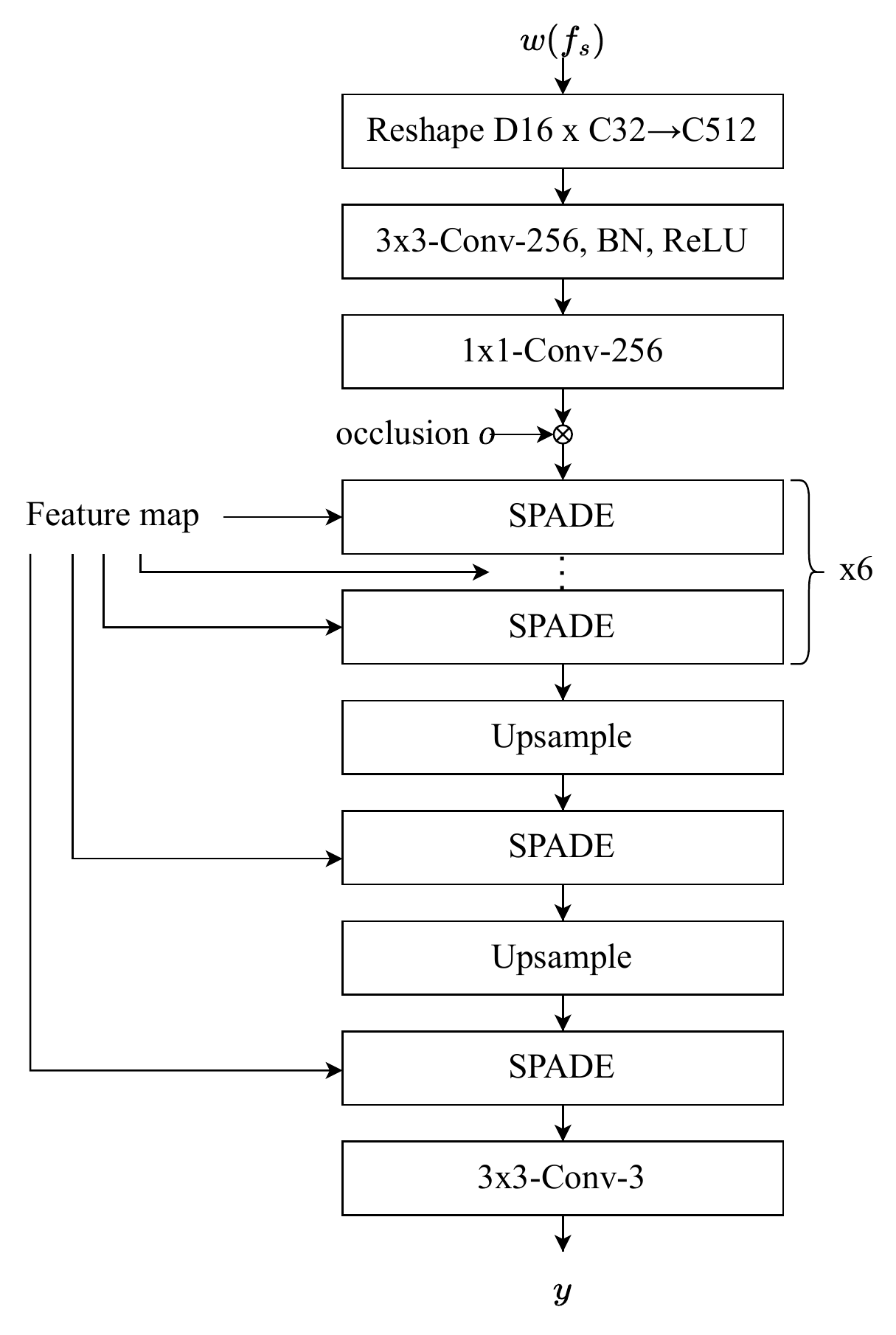}
\caption{Architectures of individual components in our model.}
\label{fig:architecture}
\vspace{-6pt} 
\end{figure*}

The implementation details of the networks are shown in Fig~\ref{fig:architecture} and described below.

\paragraph{3D Appearance feature extractor $F$} The network extracts 3D appearance feature from the source image. We use the same appearance feature extraction network proposed by Face-vid2vid~\cite{wang2021one}. It is equipped with a number of downsampling blocks, followed by a convolution layer that projects the input 2D features to 3D features. Some 3D residual blocks are then appended to compute the final 3D feature $f_s$.

\paragraph{Motion field estimator $M$.} We use the same dense motion network proposed by Face-vid2vid~\cite{wang2021one} to predict an occlusion map and the warping map. For each keypoint $k$, we generate a warping flow map $w_k$  via the first-order approximation~\cite{siarohin2019first}. Let $p_d$ denote a 3D coordinate in the feature volume for the driving image $d$. The flow field of the $k$th keypoint maps $p_d$ to $p_s$ (a 3D coordinate in the 3D feature volume of the source image $s$), via:
$$
w_k: R_s R_d^{-1}\left(p_d-x_{d, k}\right)+x_{s, k} \mapsto p_s .
$$
$w_k$ can be used to warp the source feature $f_s$ to construct a candidate warped volume, $w_k\left(f_s\right)$. For all the keypoints, we concatenate the $w_k\left(f_s\right)$ together and fed to a 3D U-Net and a softmax function to obtain $K$ 3D masks, $\left\{m_1, m_2, \ldots, m_K\right\}$. These maps satisfy the constraints that $\sum_k m_k\left(p_d\right)=1$ and $0 \leqslant m_k\left(p_d\right) \leqslant 1$ for all $p_d$. To construct the final warping map, these $K$ masks are then linearly combined with $w_k$ 's, i.e.  $ w = \sum_{k=1}^K m_k\left(p_d\right) w_k\left(p_d\right)$. To take the occlusion information into account, a $2 \mathrm{D}$ occlusion mask $o$ is predicted as well. 

\paragraph{Generator} The generator first project the warped 3D appearance feature $w(f_s)$ to 2D. The feature map is multiplied with the occlusion mask $o$. Finally, a decoder based on a series of SPADE~\cite{wang2019few} layers is applied to obtain the output image. To compare our methods with the baseline Face-vid2vid in a fair way we also apply the SPADE-based generator to the Face-vid2vid  baseline.

\subsection{Optimization} 

We adopt the ADAM optimizer with $\beta_1 = 0.5$ and $\beta_1 = 0.999$. The learning rate is set to 0.0002. We use synchronized BatchNorm for the generator. All the models are trained using 8 NVIDIA A100 GPUs for 300 epochs, which takes about 2 weeks. we check the training loss to confirm that the models converge.

\subsection{Keypoints selection process}

When faced with limited landmarks, we prioritize the eye and mouth regions while sparsely sampling the nose, eyebrow, and contour.  
For example, when total budget is only 5 landmarks, our Tab.~\ref{tab:kp_samp} shows that allocating landmarks for the mouth area is crucial for overall performance.
Additionally, our Fig.~\ref{fig:recon_sub} indicates that allocating resources to the expression code rather than landmarks is more effective when the allowed bandwidth is limited. For more than 16 points, evenly sampling the same number of landmarks per face part works well.

\begin{table}[t]
    \small
    \centering
    \tabcolsep=0.09cm
    \begin{tabular}{l|cccc|cc}
         & Concur & Eye & Nose & Mouth &  AKD$\downarrow$ & AKD-M$\downarrow$  \\
         \midrule
          Sampling strategy 1 & 1 & 2 & 2 & 0 &  1.51& 1.57\\
         Sampling strategy 2 & 0 & 2 & 1 & 2 & 1.44 & 1.35\\
         \bottomrule
    \end{tabular}
    \caption{Quantitative comparison of different landmarks sampling strategy for our w/o $e_d$.}
    \label{tab:kp_samp}
    \vspace{-15pt}
\end{table}

\subsection{Computational costs}

Our method uses a pre-extracted 3D prior to avoid repeated regeneration during training.  Using neural networks to encode images and obtain 3D priors, the DECA model incurs minimal additional inference time. Our method's inference time is comparable to existing baseline methods. For example, our approach processes a 128-frame video in only 0.5 seconds to extract face priors and keypoints, with an additional 26 seconds for our generator to synthesize all output frames. In contrast, Face-v2v requires 25 seconds to generate all output frames under the same conditions.

\section{Additional Results}
\label{app:results}

\subsection{Additional Visual Examples}
\label{app:visexample}

\begin{figure*}[t]
\vspace{-4pt}
\centering
\resizebox{0.97\linewidth}{!}{
    \setlength{\fboxrule}{4pt} 
     \setlength{\fboxsep}{0cm}
    \centering
    \begin{tabular}{>{\centering\arraybackslash}m{0.16\linewidth} >{\centering\arraybackslash}m{0.16\linewidth} >{\centering\arraybackslash}m{0.16\linewidth} >{\centering\arraybackslash}m{0.16\linewidth} >{\centering\arraybackslash}m{0.16\linewidth} }
         Source &    FOMM &   Face-vid2vid &   Ours &  Driving \\
        \includegraphics[ height=1.107\linewidth]{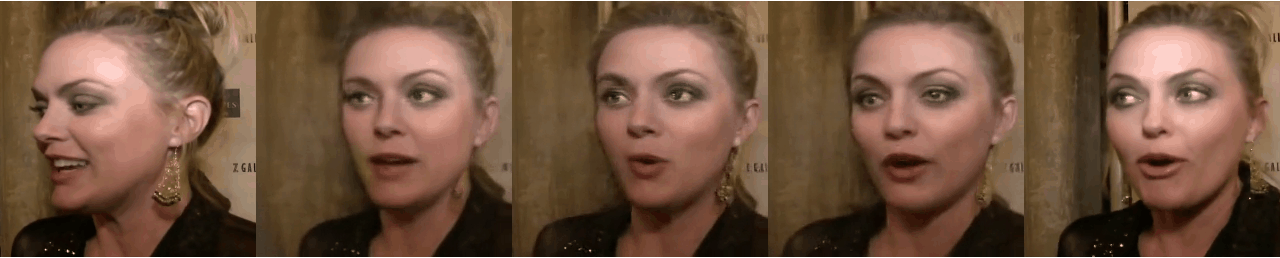}\\
        \includegraphics[ height=1.1\linewidth]{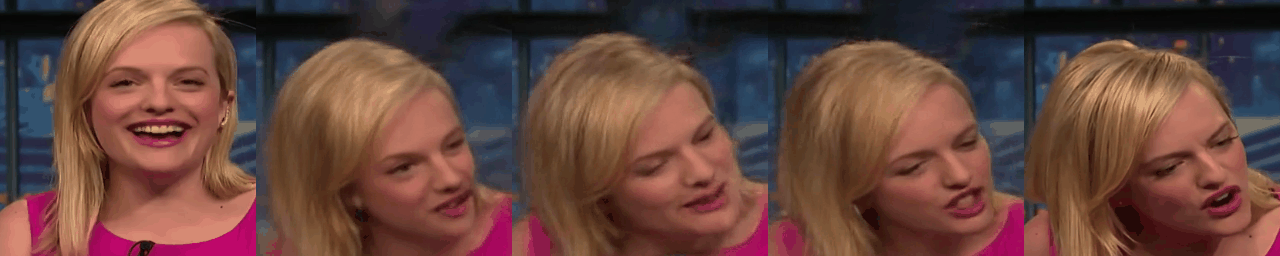}\\
       \includegraphics[ height=1.1\linewidth]{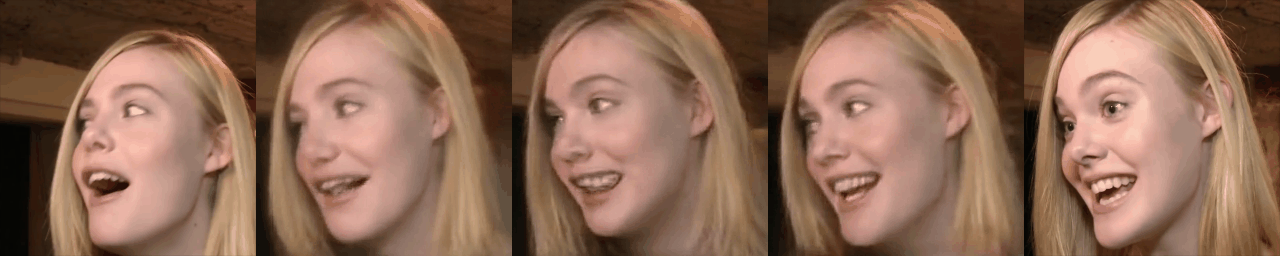}\\
               \includegraphics[ height=1.1\linewidth]{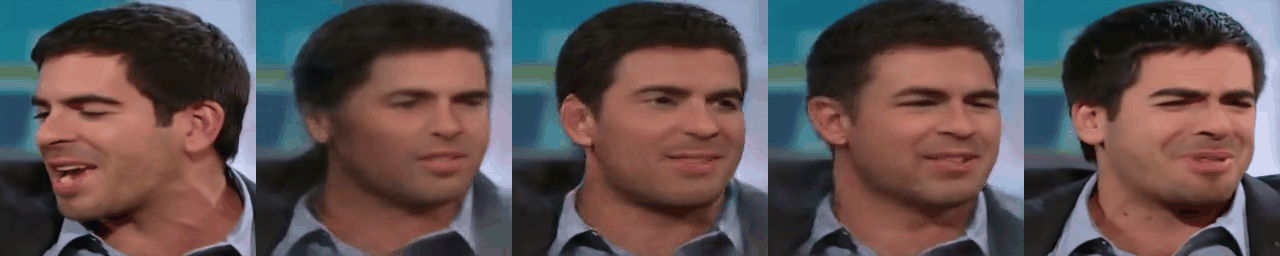}\\
     \end{tabular}  
}
\caption{Qualitative examples of reconstruction. Compared with the baselines, our method generates more realistic outputs especially when the head poses are significantly different in the source and driving frames. It also preserves the emotion in the original frame better.}
\label{fig:add_qualitative_Recons}
\vspace{-6pt} 
\end{figure*}

\begin{figure*}[t]
\vspace{-4pt}
\centering
\resizebox{0.97\linewidth}{!}{
    \setlength{\fboxrule}{4pt} 
     \setlength{\fboxsep}{0cm}
    \centering
    \begin{tabular}{>{\centering\arraybackslash}m{0.16\linewidth} >{\centering\arraybackslash}m{0.16\linewidth} >{\centering\arraybackslash}m{0.16\linewidth} >{\centering\arraybackslash}m{0.16\linewidth} >{\centering\arraybackslash}m{0.16\linewidth} }
         Source &    FOMM &   Face-vid2vid &   Ours &  Driving \\
         \includegraphics[ height=1.1\linewidth]{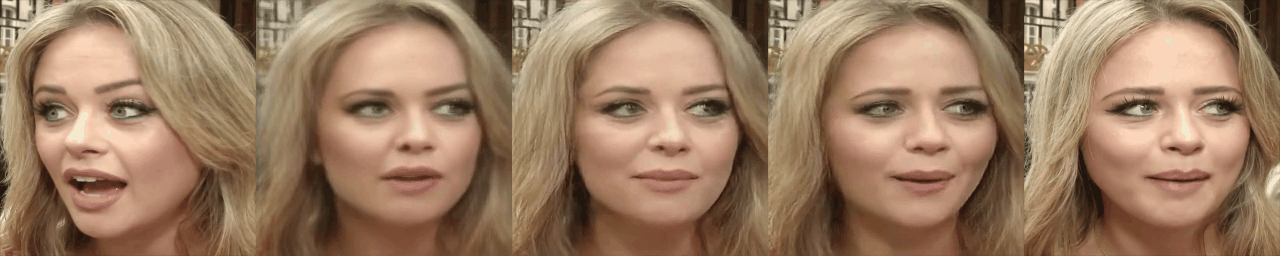}\\
      \includegraphics[ height=1.1\linewidth]{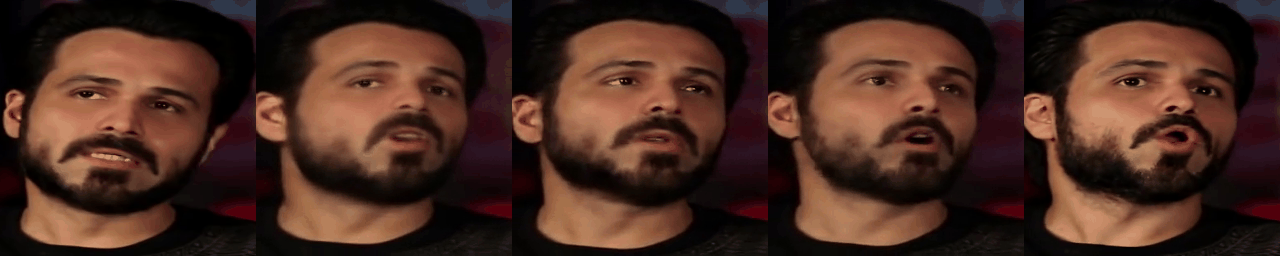}\\
        \includegraphics[ height=1.1\linewidth]{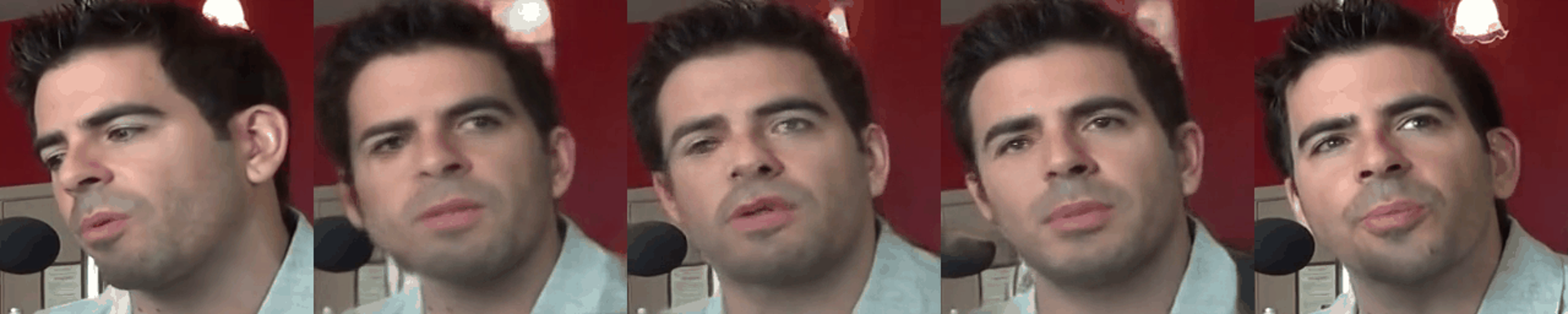}\\
        \includegraphics[ height=1.1\linewidth]{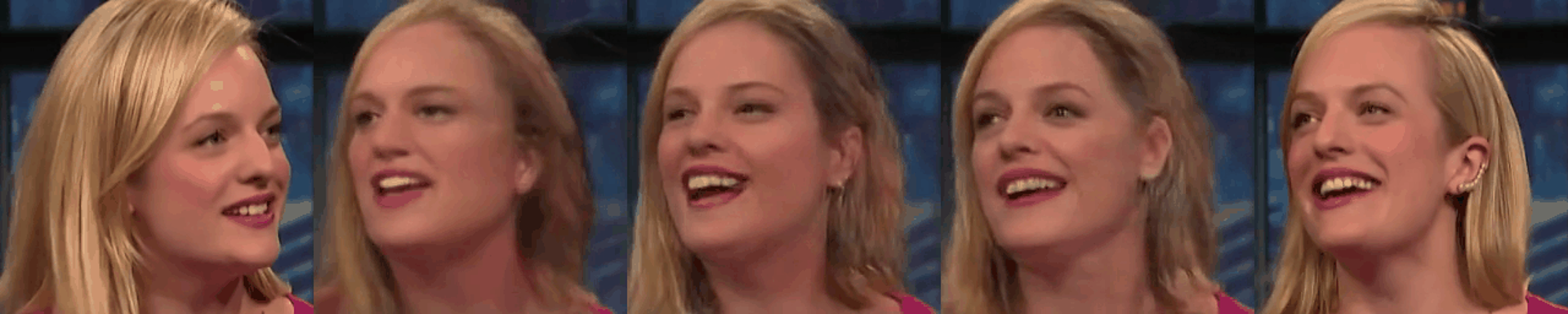}\\
     \end{tabular}  
}
\caption{Qualitative examples of reconstruction. Compared with the baselines, our method generates more realistic outputs especially when the head poses are significantly different in the source and driving frames. It also preserves the emotion in the original frame better.}
\label{fig:add_qualitative_Recons_2}
\vspace{-6pt} 
\end{figure*}

\paragraph{Reconstruction} In Figure~\ref{fig:add_qualitative_Recons} and Figure~\ref{fig:add_qualitative_Recons_2}, we provide additional visual examples of video reconstruction. In the folder \textbf{recons}, we provide video examples for reconstruction. 

\begin{figure*}[t]
\vspace{-4pt}
\centering
\resizebox{0.97\linewidth}{!}{
    \setlength{\fboxrule}{4pt} 
     \setlength{\fboxsep}{0cm}
    \centering
    \begin{tabular}{>{\centering\arraybackslash}m{0.16\linewidth} >{\centering\arraybackslash}m{0.16\linewidth} >{\centering\arraybackslash}m{0.16\linewidth} >{\centering\arraybackslash}m{0.16\linewidth} >{\centering\arraybackslash}m{0.16\linewidth} }
         Source &    FOMM &   Face-vid2vid &   Ours &  Driving \\
        \includegraphics[ height=1.1\linewidth]{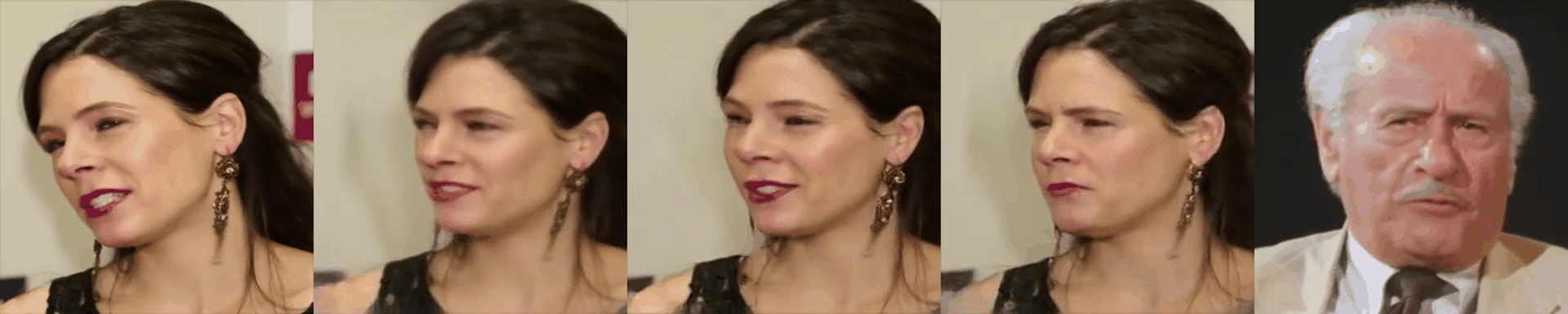}\\
        \includegraphics[ height=1.1\linewidth]{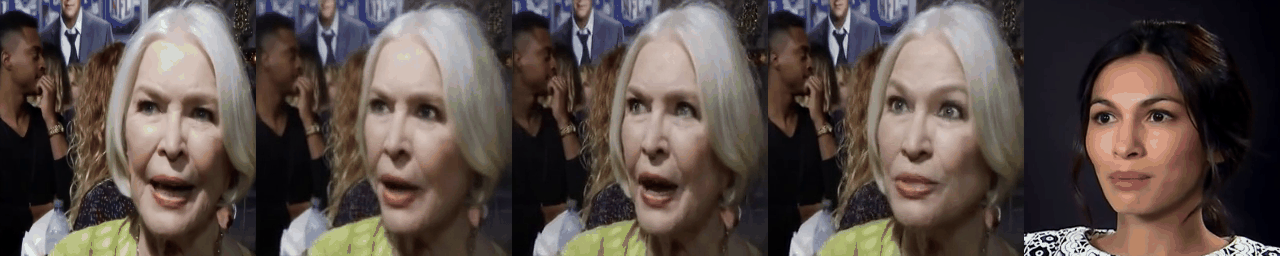}\\
        \includegraphics[ height=1.1\linewidth]{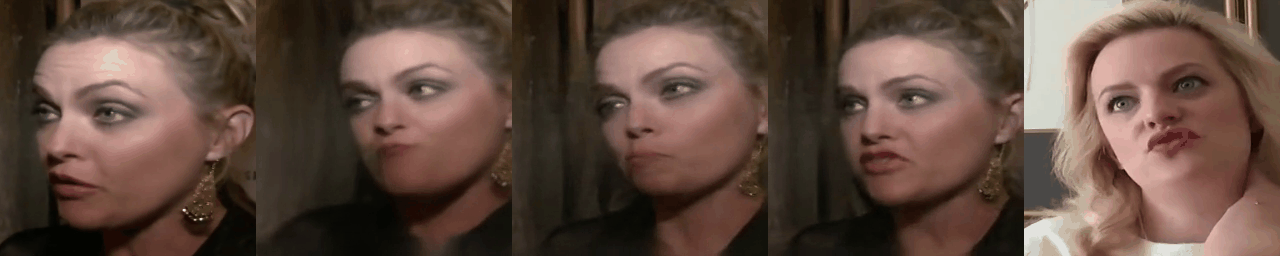}\\ 
        \includegraphics[ height=1.1\linewidth]{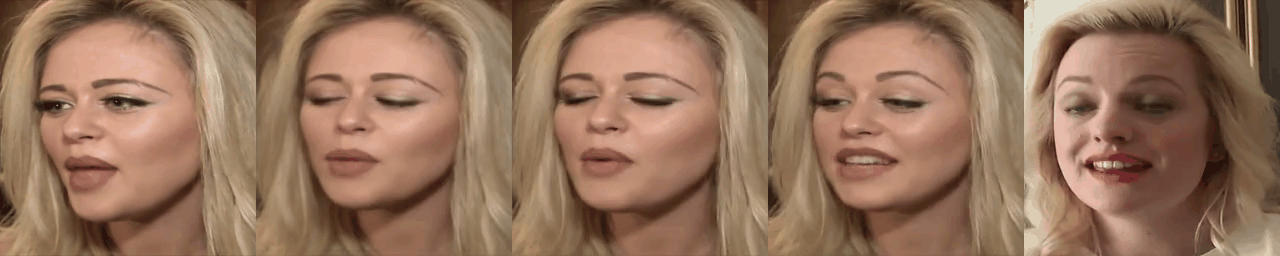}\\ 
         \includegraphics[ height=1.1\linewidth]{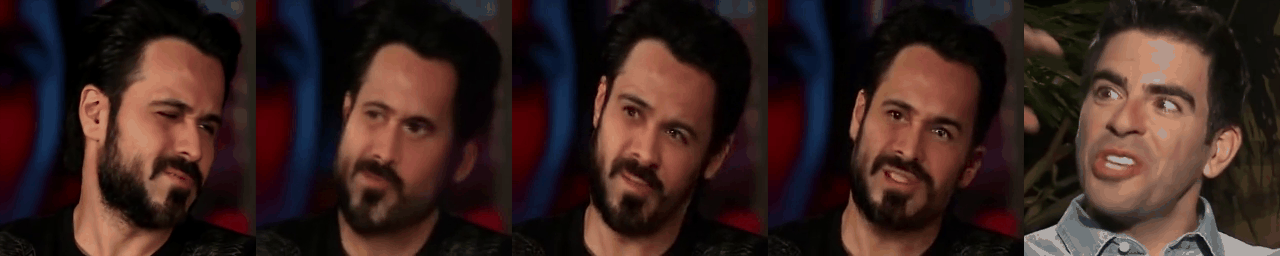}\\ 
     \end{tabular}  
}
\caption{Qualitative examples of re-enactment. For a fair comparison with FOMM~\cite{siarohin2019first}, the visual examples for re-enactment are reported via the relative motion adopted in FOMM. Compared with the baselines, our method generates more realistic outputs especially when the head poses are significantly different in the source and driving frames. Also, it transfers fine-grained expression more faithfully. }
\label{fig:add_qualitative_trans}
\vspace{-6pt} 
\end{figure*}

\paragraph{Re-enactment} In Figure~\ref{fig:add_qualitative_trans}, we provide additional visual examples of cross-person re-enactment. In the folder \textbf{re-enactment}, we provide video examples for re-enactment. 

\begin{figure*}[t]
\vspace{-4pt}
\centering
\resizebox{0.97\linewidth}{!}{
    \setlength{\fboxrule}{4pt} 
     \setlength{\fboxsep}{0cm}
    \centering
    \begin{tabular}{>{\centering\arraybackslash}m{0.16\linewidth} >{\centering\arraybackslash}m{0.16\linewidth} >{\centering\arraybackslash}m{0.16\linewidth} >{\centering\arraybackslash}m{0.16\linewidth} >{\centering\arraybackslash}m{0.16\linewidth} }
         Source &    Driving &   Unmodified &   Neutral &  Happy \\
        \includegraphics[ height=1.3\linewidth]{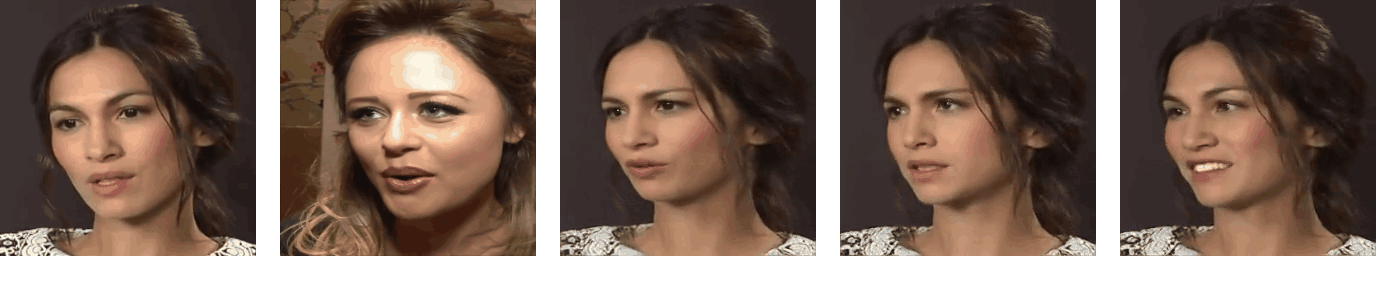}\\
        \includegraphics[ height=1.3\linewidth]{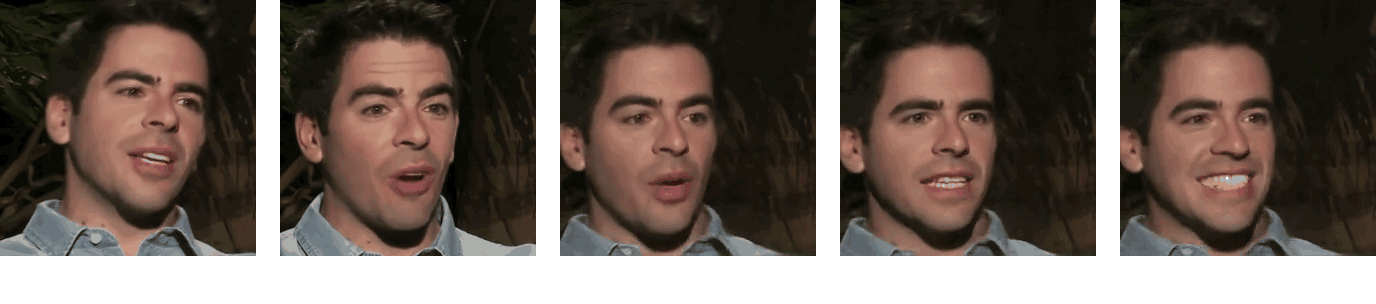}\\ 
     \end{tabular}  
}
\caption{Additional facial expression manipulation results. Our method can simultaneously transfer the coarse head motion from the driving frame and alter the fine-grained expression using the expression feature. Note that the results are obtained by plugging in an off-the-shelf emotion translation model during inference without any change to the model. }
\label{fig:add_qualitative_emo}
\vspace{-6pt} 
\end{figure*}

\paragraph{Emotion manipulation} In Figure~\ref{fig:add_qualitative_emo}, we provide additional visual examples of emotion manipulation. In the folder \textbf{emotion}, we provide video examples for emotion manipulation.

\begin{figure*}[t]
\vspace{-4pt}
\centering
\resizebox{0.97\linewidth}{!}{
    \setlength{\fboxrule}{4pt} 
     \setlength{\fboxsep}{0cm}
    \centering
    \begin{tabular}{>{\centering\arraybackslash}m{0.12\linewidth} >{\centering\arraybackslash}m{0.12\linewidth} >{\centering\arraybackslash}m{0.12\linewidth} >{\centering\arraybackslash}m{0.12\linewidth} >{\centering\arraybackslash}m{0.12\linewidth} 
    >{\centering\arraybackslash}m{0.12\linewidth} 
    >{\centering\arraybackslash}m{0.12\linewidth} 
    }
         Source &    Driving &   Driving Pose &   Pose 1 &  Pose 2 & Pose 3 & Pose 4 \\
        \includegraphics[trim={0 1.68cm 0 0}, clip, height=1.1\linewidth]{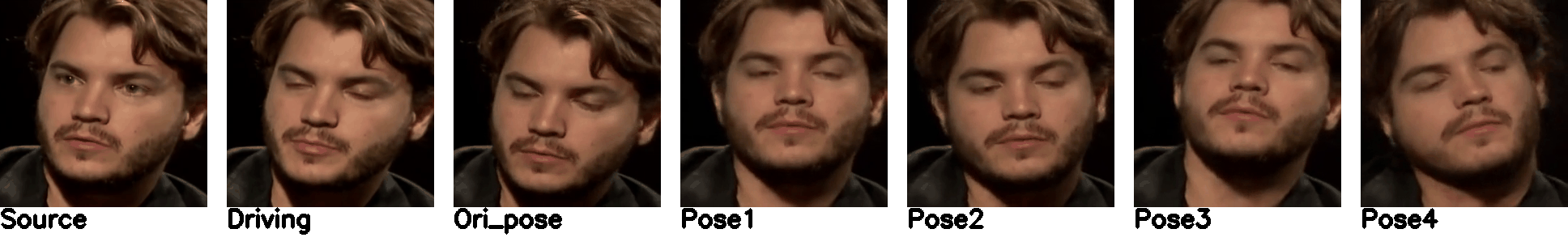}\\
        \includegraphics[trim={0 1.68cm 0 0}, clip, height=1.1\linewidth]{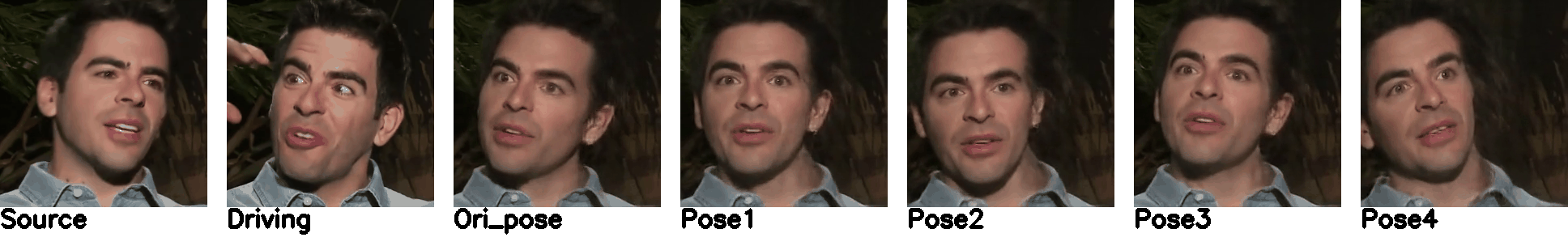}\\
        \includegraphics[trim={0 1.68cm 0 0}, clip, height=1.1\linewidth]{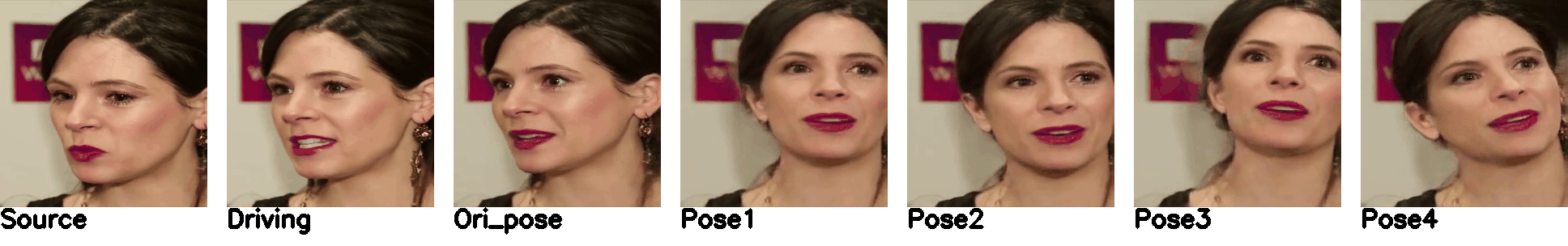}\\
     \end{tabular}  
}
\caption{Facial pose manipulation results. Our method can simultaneously transfer the fine-grained expression from the driving frame, and preserve or alter the coarse head motion by modifying the pose code to render new keypoints. }
\label{fig:add_qualitative_pose}
\vspace{-6pt} 
\end{figure*}

\paragraph{Pose manipulation} In Figure~\ref{fig:add_qualitative_pose}, we provide additional visual examples of pose manipulation.

\subsection{Additional Quantitative Results}
\label{app:quanres}

We provide additional quantitative results on the reconstruction accuracy versus bitrate as mentioned in Figure~\ref{fig:recon_sub} in the main paper. 

\begin{figure*}
    \centering
    \includegraphics[width=0.6\textwidth]{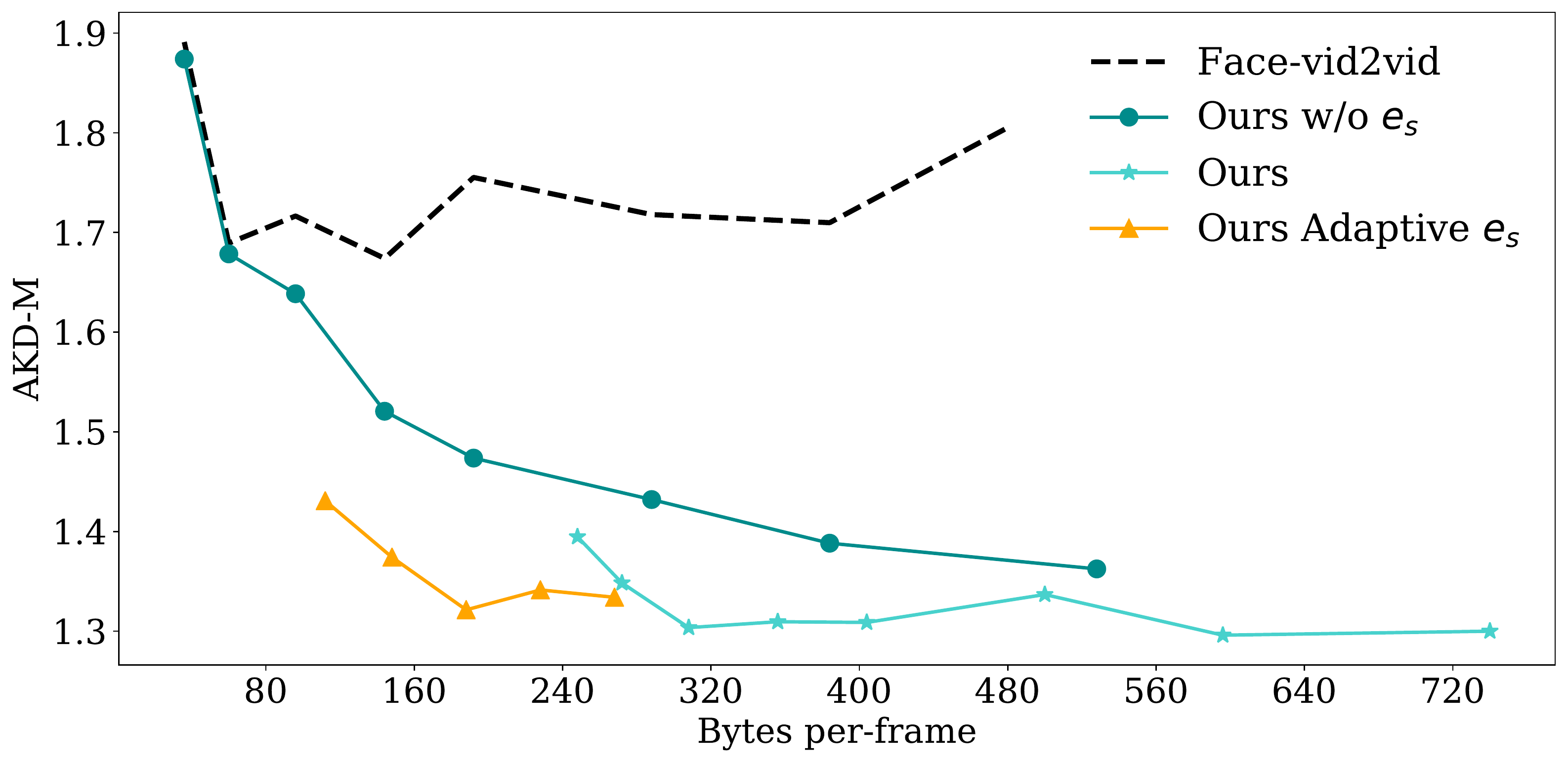}
    \includegraphics[width=0.6\textwidth]{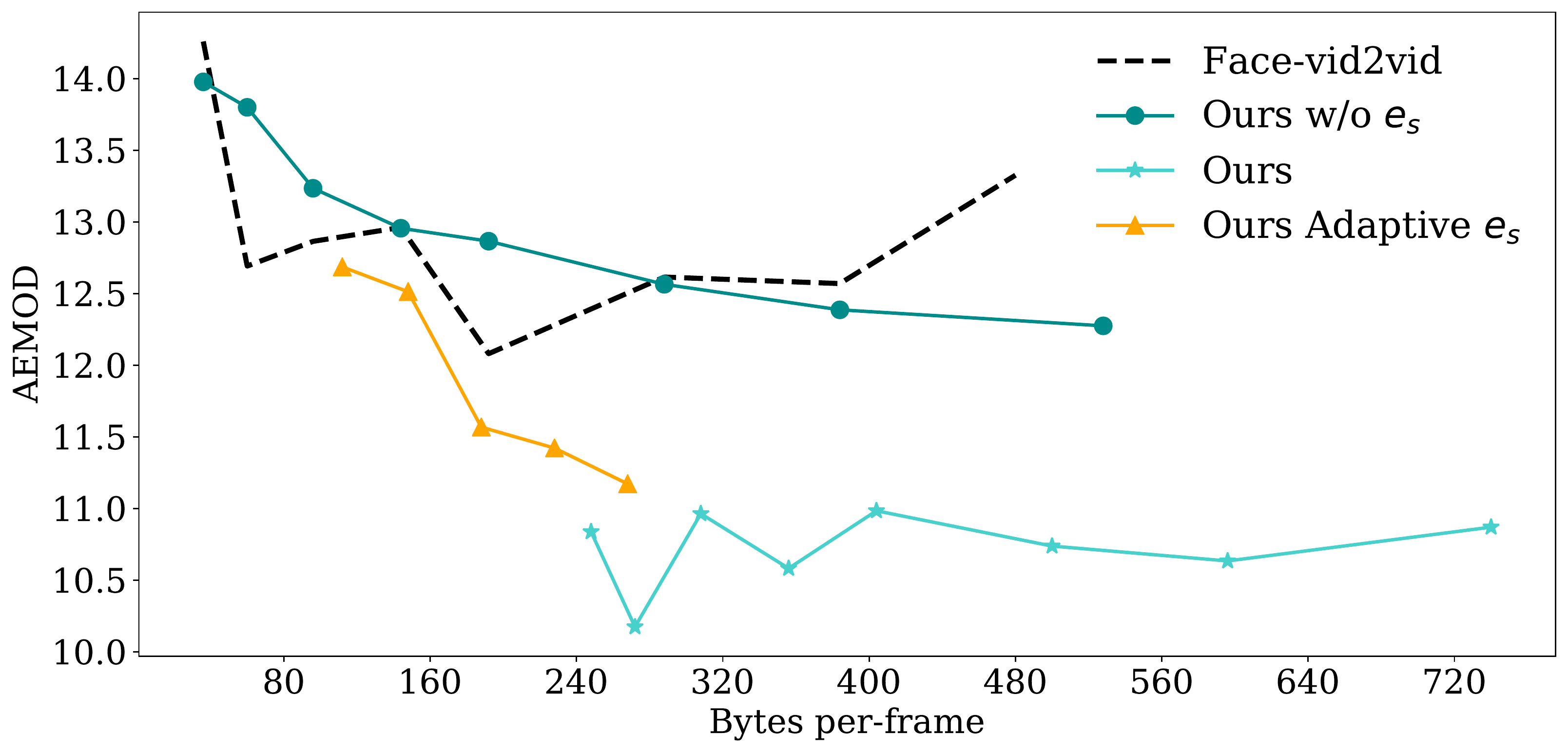}
    \caption{Other metrics corresponds to Figure~\ref{fig:recon_sub}. Reconstruction accuracy versus bitrate on the hard subset. The driving video can be encoded using the keypoints and expression features, and we control the bitrate by changing the number of keypoints and size of $e_d$ (for Ours adaptive $e_d$). Our method achieves the best reconstruction accuracy under all bitrates.}
    \label{fig:append_subset_bitrate}
\end{figure*}

In Figure~\ref{fig:append_subset_bitrate}, we provide the AKD-M and AEMOD for the hard subset. The results show that our model consistently outperforms the unsupervised keypoints-based method across different bitrate and evaluation metrics.

\begin{figure*}
    \centering
    \includegraphics[width=0.6\textwidth]{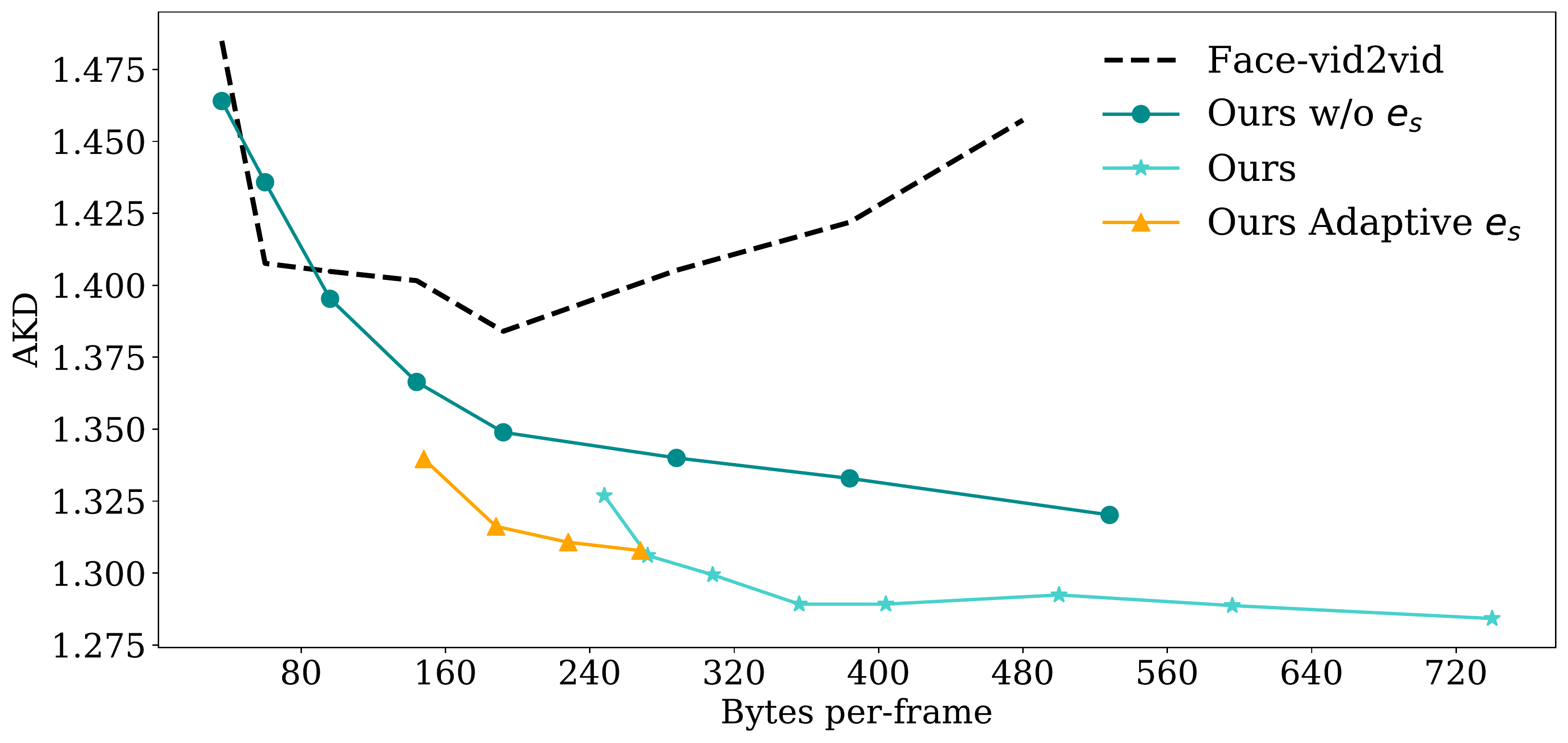}
    \includegraphics[width=0.6\textwidth]{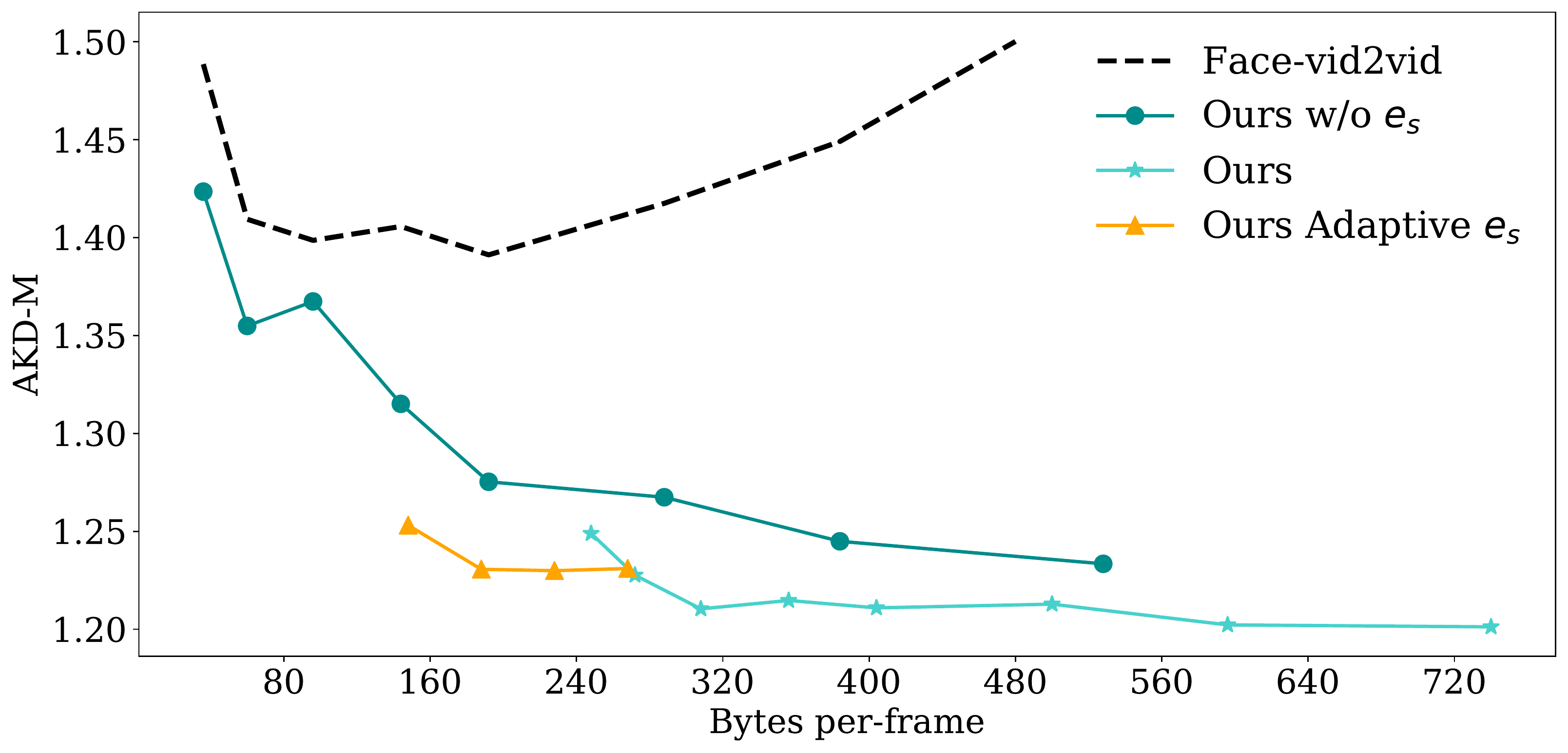}
    \includegraphics[width=0.6\textwidth]{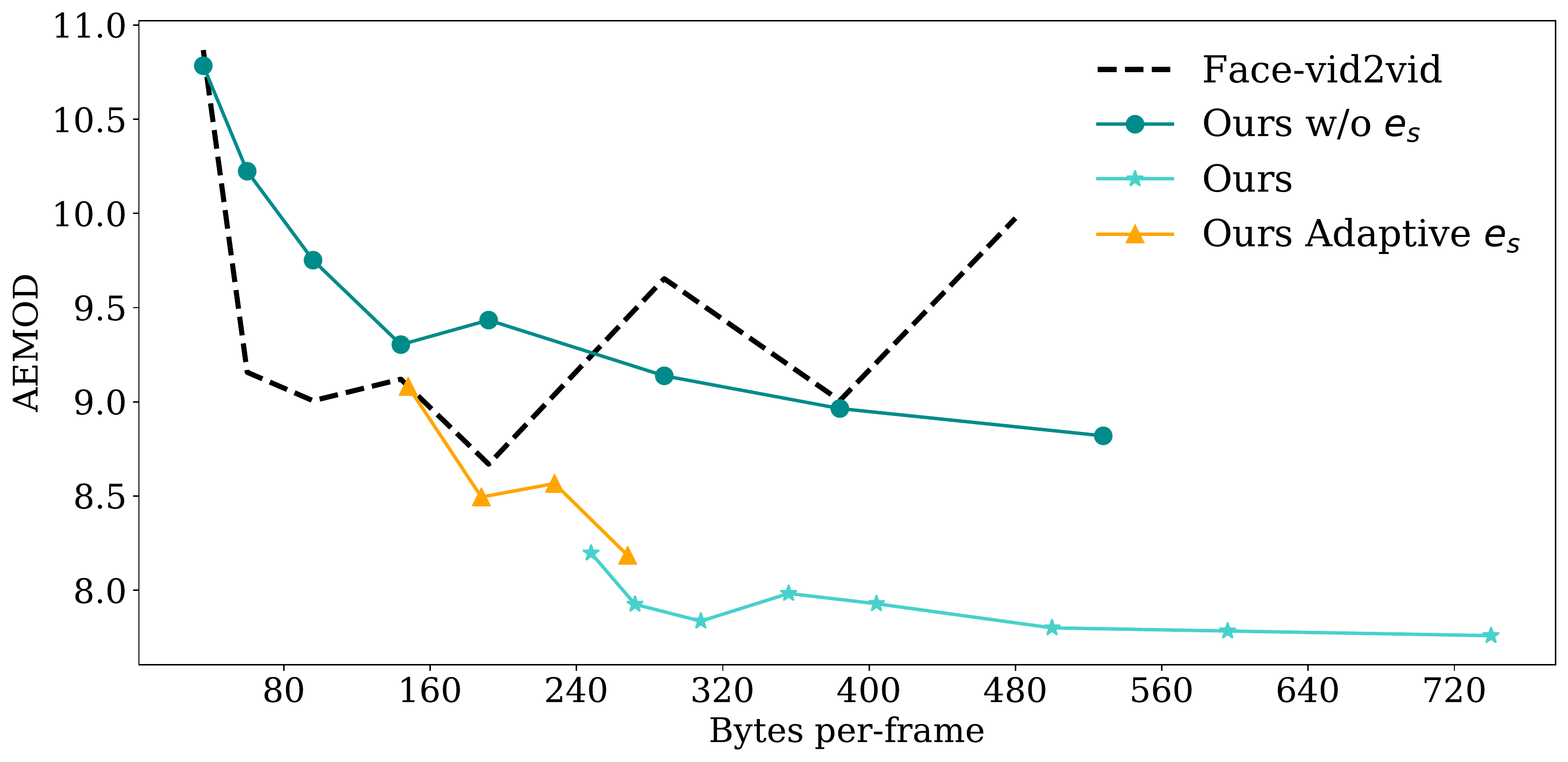}
    \caption{Reconstruction accuracy versus bitrate for the full testset. The driving video can be encoded using the keypoints and expression features, and we control the bitrate by changing the number of keypoints and size of $e_d$ (for Ours adaptive $e_d$). Our method achieves the best reconstruction accuracy under most of the bitrates. 
    }
    \label{fig:append_full_bitrate}
\end{figure*}

In Figure~\ref{fig:append_full_bitrate}, we provide the AKD, AKD-M, and AEMOD for the full testset. The results again validate the superiority of our method.

\end{document}